\documentclass[10pt,twocolumn,letterpaper]{article}

\usepackage{cvpr}              

\usepackage[T1]{fontenc}

\usepackage{cuted} 

\usepackage{currfile} 

\usepackage{caption} 
\definecolor{cvprblue}{rgb}{0.21,0.49,0.74}
\usepackage[pagebackref,breaklinks,colorlinks,allcolors=cvprblue]{hyperref}

\def\paperID{28260} 
\def\confName{CVPR}
\def\confYear{2027}

\title{Family-Constrained Metamers Reveal Symmetric and Antisymmetric Organization of Visual Encoding Models}

\author{%
  Hayato Ono\\
  Graduate School of Information Science and Technology\\
  The University of Tokyo, Tokyo, Japan\\
  {\small\texttt{hayato-0628@g.ecc.u-tokyo.ac.jp}}
  \and
  Ayumu Yamashita\\
  Graduate School of System Informatics\\
  Kobe University, Hyogo, Japan\\
  {\small\texttt{ayumu722@people.kobe-u.ac.jp}}
  \and
  Masataka Sawayama\\
  1. Graduate School of Information Science and Technology\\
  Hokkaido University, Hokkaido, Japan\\
  2. Prometech CG Research, Tokyo, Japan\\
  {\small\texttt{masa.sawayama@gmail.com}}
  \and
  Kaoru Amano\\
  Graduate School of Information Science and Technology\\
  The University of Tokyo, Tokyo, Japan\\
  {\small\texttt{kaoru\_amano@ipc.i.u-tokyo.ac.jp}}
  \and
  Tomoya Nakai\\
  Graduate School of Information Science and Technology\\
  The University of Tokyo, Tokyo, Japan\\
  {\small\texttt{tnakai@g.ecc.u-tokyo.ac.jp}}%
}

\begin{document}
\maketitle
\begin{abstract}

Visual encoding models are typically compared by their neural-response prediction accuracy, which does not reveal whether image changes preserving one model's predicted responses also preserve another's. We examine this directional correspondence using family-constrained metamers jointly optimized to preserve the predicted responses of encoding models within each architecture family. Using the Natural Scenes Dataset, we fit 1,280 voxel-wise encoding models spanning 80 pretrained backbones, 15 architecture families, eight subjects, and V1 and hV4. We synthesize 9,000 metamers from models fitted to three subjects and evaluate transfer with models from all eight subjects. Cross-family transfer profiles are highly reliable across target subsets (mean Spearman--Brown-corrected split-half reliability 0.977) and show strong correspondence when the evaluating subject changes (mean Pearson correlation 0.872). Family-wide transfer levels and generator--evaluator roles account for most symmetric and antisymmetric transfer structure, respectively, while pair-specific residuals remain reproducible. In within-subject evaluations, these residuals reduce held-out-target prediction MSE by 6.4\% relative to family-wide effects alone. These findings reveal stable family-wide tendencies and pair-specific response-preserving relationships in fitted visual encoding models.

\end{abstract}
    
\section{Introduction}
\label{sec:introduction}

\begin{figure*}[t]
    \centering
    \includegraphics[width=\textwidth]{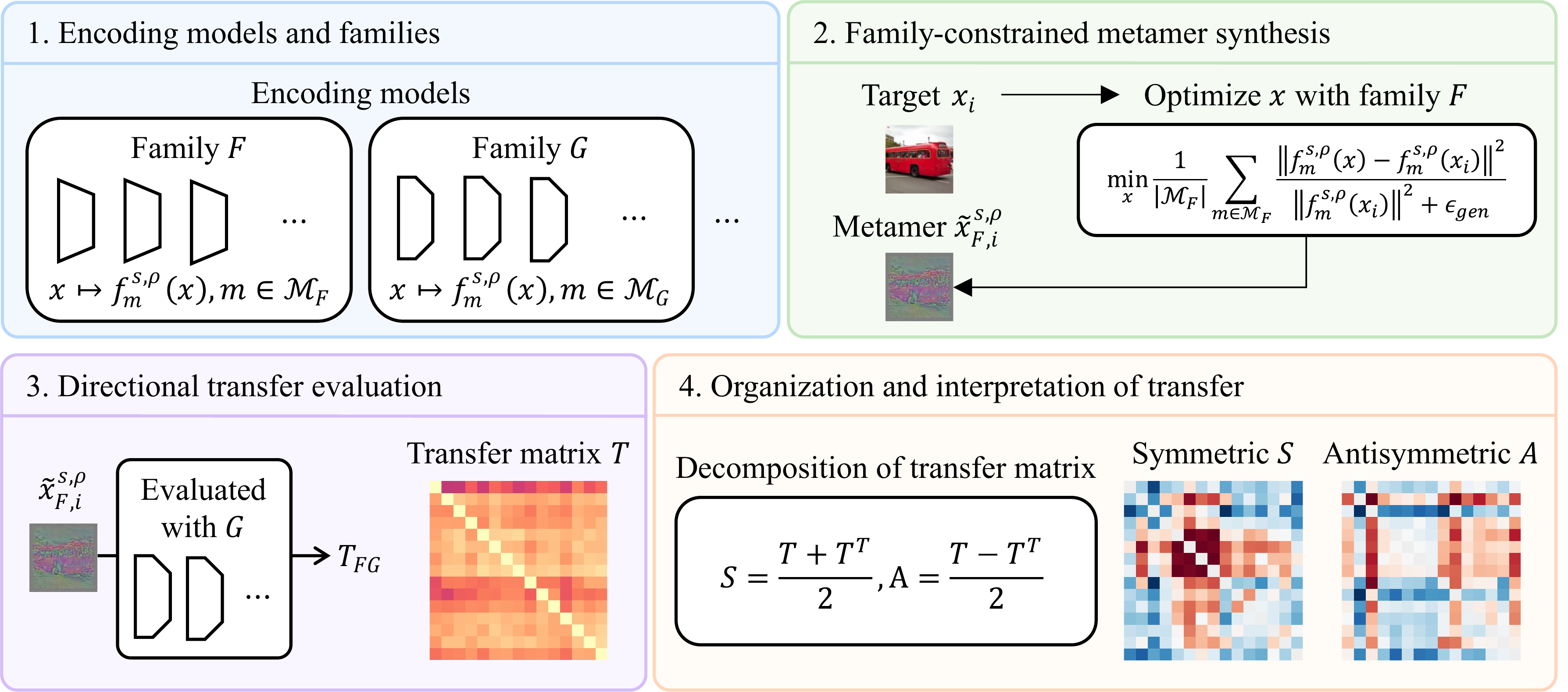}
    \caption{\textbf{Overview of the study.}
    (1) Encoding models are fitted to NSD responses and grouped into architecture families. (2) Family-constrained synthesis generates metamers that approximately preserve each constituent model's predicted responses. (3) Cross-family evaluation produces a directional transfer matrix $T$. (4) The transfer matrix $T$ is analyzed through symmetric and antisymmetric decomposition, together with family-wide generating and evaluating effects and pair-specific residuals.}
    \label{fig:overview}
\end{figure*}

Visual encoding models provide a quantitative framework for relating visual representations to cortical activity by predicting neural responses to images~\cite{kay2008identifying,naselaris2011encoding}. Deep neural network features have been used to predict visual-cortical responses in nonhuman primates~\cite{yamins2014performance-optimized} and humans~\cite{guclu2015deep_ab,eickenberg2017seeing}. Large-scale datasets such as the Natural Scenes Dataset (NSD)~\cite{allen2022a} enable subject-specific encoding models to be fitted and evaluated using responses to thousands of natural images, allowing encoding models built on diverse feature representations to be compared using prediction accuracy.

However, prediction accuracy alone does not reveal which image changes leave an encoding model's predicted cortical responses approximately unchanged. These response-preserving properties characterize which distinctions between visual stimuli are not resolved by the model's predictions. Metamer synthesis provides a way to probe these properties by generating images that differ from a target while approximately matching its representation under a specified model~\cite{freeman2011metamers,feather2019metamers,feather2023model}. This approach has also been applied to encoding models, using model predictions to synthesize stimuli targeting recorded neural responses~\cite{cobos2022it}. Because encoding models based on different feature representations can be fitted to the same cortical response data, their response-preserving properties can be compared in a common neural prediction space.

A metamer synthesized to preserve one model's predicted response to a target can be evaluated with another model to test whether the evaluating model's predicted response is also preserved. This comparison, termed metamer transfer, is directional: reversing the generating and evaluating roles requires a separate synthesis and may yield a different score. Previous studies have examined such cross-model response preservation~\cite{feather2019metamers,feather2023model,nanda2022measuring}. However, pairwise transfer scores alone cannot distinguish general generating and evaluating tendencies from relationships specific to individual model pairs. To examine this structure beyond individual model variants, we group encoding models by backbone architecture and systematically compare transfer across families using metamers constrained by their constituent models. Our central question is how much of directional transfer is captured by family-wide transfer levels and generator–evaluator roles, and whether the remaining pair-specific structure is reproducible across target images and improves prediction on held-out targets.

To address this question, we develop a framework for characterizing directional transfer among visual encoding-model families using family-constrained metamers (Fig.~\ref{fig:overview}). Using NSD, we fit subject-specific voxel-wise encoding models of V1 and hV4 based on 80 pretrained vision backbones grouped into 15 architecture families. For each target image, we jointly optimize a metamer image per generating family to approximately preserve each constituent encoding model's predicted response to that target. We then evaluate each metamer using encoding models from all families to construct a directional family-by-family transfer matrix. We generate 9,000 metamer images using models fitted to three subjects and evaluate them using models fitted to all eight NSD subjects, enabling both within-subject and cross-subject evaluations while keeping the region-of-interest fixed. We decompose the resulting transfer matrices into symmetric and antisymmetric components and separate family-wide transfer levels and generator--evaluator roles from pair-specific residuals.

Cross-family transfer is reproducible across target subsets and largely preserved across evaluating subjects when metamers are held fixed. In within-subject evaluations, family-wide transfer levels account for an average of 82.8\% of the centered symmetric squared magnitude, while generator–evaluator roles account for 90.7\% of the pooled antisymmetric squared magnitude. Reliable pair-specific residuals remain and reduce mean squared error in transfer prediction on held-out targets by 6.4\% relative to family-wide effects alone under five-fold target cross-validation. These findings identify reproducible relationships with predictive value in fitted visual encoding models.
\section{Related work}
\label{sec:related_work}

\subsection{Encoding Models of Visual Cortex}

Visual encoding models have used both hand-designed visual features~\cite{kay2008identifying,naselaris2009bayesian,naselaris2011encoding} and representations from task-optimized deep neural networks~\cite{yamins2014performance-optimized,guclu2015deep_ab,eickenberg2017seeing,st-yves2018the,wen2018neural}. Large-scale datasets such as NSD~\cite{allen2022a} allow encoding models based on different feature representations to be fitted to the same cortical response data and compared using prediction accuracy. While such comparisons characterize predictive performance, our study examines which image changes preserve the cortical responses predicted by fitted encoding models, including their pretrained feature extractors and learned voxel-wise readouts.

\subsection{Metamer Synthesis and Response Preservation}

Perceptual metamer studies test whether images matched under a visual model are indistinguishable to human observers~\cite{freeman2011metamers,wallis2016testing,wallis2019image}. Neural-network metamer studies extend this approach to learned representations, revealing discrepancies between model-defined response preservation and human recognition that conventional neural prediction benchmarks may miss~\cite{feather2019metamers,feather2023model}. Related work has used encoding models to synthesize stimuli targeting recorded neural responses~\cite{cobos2022it} and digital twins to probe single-neuron response properties in mouse visual cortex~\cite{ding2026functional}. Beyond comparisons with human perception or recorded neural responses, cross-model evaluations test whether response preservation extends across models, while ensemble synthesis constrains multiple models~\cite{feather2023model,boehm2026understanding}. Building on encoding-model synthesis and cross-model evaluation, our study characterizes response-preserving transfer across human visual encoding-model families using jointly constrained metamers.

\subsection{Representational Similarity and Directional Transfer}

Representational similarity analysis (RSA), centered kernel alignment (CKA), canonical-correlation-based methods, and generalized shape metrics compare representations across brains and models~\cite{kriegeskorte2008representational,khaligh-razavi2014deep,kornblith2019similarity,raghu2017svcca,morcos2018insights,williams2021generalized}. Voxel-wise encoding RSA reveals increased similarity among cortical-response-fitted models~\cite{conwell2024a}, while equivariance and model-stitching analyses examine transformation-related properties and functional compatibility~\cite{lenc2015understanding,bansal2021revisiting}. However, these approaches do not directly test whether image changes preserving one fitted encoding model's predicted responses also preserve another's. STIR explicitly examines directional shared invariances between network representations~\cite{nanda2022measuring}. We extend this perspective to encoding-model families fitted to cortical responses, decomposing directional metamer transfer into symmetric and antisymmetric components, separating family-wide transfer levels and generator–evaluator roles from pair-specific residuals, and assessing their reproducibility and predictive value on held-out targets.

\section{Method}
\label{sec:method}

In this section, we describe a framework for characterizing response-preserving transfer among subject- and region-of-interest (ROI)-specific encoding-model families. The framework comprises encoding-model fitting, family-constrained metamer synthesis, directional transfer evaluation, and decomposition into symmetric and antisymmetric components, family-wide effects, and pair-specific residuals.

\subsection{Encoding Model Fitting}
\label{sec:encoding_model_fitting}

For each subject $s$ and ROI $\rho$, we fit voxel-wise encoding models that map image features to neural responses~\cite{kay2008identifying,naselaris2011encoding}. Each model consists of a pretrained visual feature extractor, principal component analysis (PCA), and a voxel-wise ridge-regression readout~\cite{guclu2015deep_ab,st-yves2018the,wen2018neural}. We denote the raw-response prediction of model $m$ for image $x$ by $f_m^{s,\rho}(x)$.

Images are split into training, validation, and test sets by canonical image identity, with repeated responses averaged per image. For each candidate feature layer, activations are flattened and PCA is fitted on the training set. The PCA transformation is held fixed during ridge cross-validation, which selects regularization separately for each voxel. The feature layer is selected based on validation-set Pearson correlation.

With the selected layer fixed, the encoding model is refitted on the combined training and validation data, re-estimating PCA, voxel-wise ridge regularization through cross-validation, and readout weights. The fitted response means and effective scales are retained with each final model.

\subsection{Family-Constrained Metamer Synthesis}
\label{sec:metamer_generation}

For each subject $s$ and ROI $\rho$, target images are selected from the test set based on agreement between predicted and measured voxel responses. Let $\mathbf{y}_i^{s,\rho}$ denote the measured neural response vector for image $x_i$, and let $\mathcal{F}$ denote the set of $L$ architecture families, with $\mathcal{M}_F$ containing the encoding models in family $F$. We define a family-balanced prediction-agreement score as
\begin{equation}
    a_i^{s,\rho}
    =
    \frac{1}{L}
    \sum_{F\in\mathcal{F}}
    \frac{1}{|\mathcal{M}_F|}
    \sum_{m\in\mathcal{M}_F}
    \operatorname{corr}_{v}
    \left(f_m^{s,\rho}(x_i),\mathbf{y}_i^{s,\rho}\right),
    \label{eq:target_selection}
\end{equation}
where $\operatorname{corr}_{v}$ denotes Pearson correlation across voxels. Images are ranked by this score to form high- and low-agreement target cohorts, denoted by $\kappa\in\{\mathrm{top},\mathrm{bottom}\}$, with target sets $\mathcal{I}_{\kappa}^{s,\rho}$. The same selected targets are used across all generating families.

For each selected target image $x_i$, a single metamer is optimized per generating family $F$ to approximate each constituent model's predicted response by minimizing
\begin{equation}
    \mathcal{L}_{F}^{s,\rho}(x;x_i)
    =
    \frac{1}{|\mathcal{M}_F|}
    \sum_{m\in\mathcal{M}_F}
    \frac{
        \left\|f_m^{s,\rho}(x)-f_m^{s,\rho}(x_i)\right\|^2
    }{
        \left\|f_m^{s,\rho}(x_i)\right\|^2+\epsilon_{\mathrm{gen}}
    }.
    \label{eq:family_objective}
\end{equation}

Each model's loss measures the squared Euclidean distance between raw predicted voxel-response vectors, normalized by its target-response energy. A small constant $\epsilon_{\mathrm{gen}}>0$ ensures numerical stability. Model-wise losses are averaged with equal weights, rather than averaging response vectors across models.

With encoding-model parameters fixed, image pixels are optimized directly using Adam~\cite{kingma2015adam}, yielding the metamer $\hat{x}_{F,i}^{s,\rho}$.

\subsection{Directional Metamer Transfer}
\label{sec:metamer_evaluation}

A metamer generated using models fitted to subject $s$ is evaluated with models fitted to subject $u$, keeping the ROI $\rho$ fixed. Evaluations with $s=u$ are termed source-self evaluations, whereas those with $s\ne u$ are termed cross-subject evaluations. Each evaluating model compares its predicted voxel responses for the metamer and the original target.

Unlike the synthesis loss, transfer evaluation uses fit-standardized responses. For each evaluating model $m$, let $\boldsymbol{\mu}_m$ and $\boldsymbol{\sigma}_m$ denote the vectors of per-voxel response means and effective scales estimated during final fitting. The standardized prediction is
\begin{equation}
    z_m^{u,\rho}(x)
    =
    \left(f_m^{u,\rho}(x)-\boldsymbol{\mu}_m\right)
    \oslash
    \boldsymbol{\sigma}_m,
    \label{eq:standardized_prediction}
\end{equation}
where $\oslash$ denotes element-wise division, and the dependence of the normalization parameters on $u$ and $\rho$ is implicit.

To normalize response discrepancies on a model-relative scale, we use a fixed reference set $\mathcal{C}=\{c_1,\ldots,c_K\}$ of $K=100$ images selected from the official NSD stimulus pool. These images were absent from the presentation records of all eight subjects and thus disjoint from all model-fitting, test, and synthesis-target images. The same reference set was used across all subjects, ROIs, and encoding models. The mean squared response distance between distinct reference images for model $m$ is
\begin{equation}
    b_m
    =
    \frac{1}{K(K-1)}
    \sum_{j\ne l}
    \frac{1}{V}
    \left\|z_m^{u,\rho}(c_j)-z_m^{u,\rho}(c_l)\right\|^2,
    \label{eq:reference_distance}
\end{equation}
where $V$ is the number of evaluated voxels, and the sum includes all $K(K-1)$ ordered pairs of distinct reference images. The squared norm sums over voxels, and division by $V$ yields their mean squared difference. The model-level transfer error is
\begin{equation}
    e_{i,F,m}^{s\to u,\rho}
    =
    \frac{
        \frac{1}{V}
        \left\|z_m^{u,\rho}(\hat{x}_{F,i}^{s,\rho})-z_m^{u,\rho}(x_i)\right\|^2
    }{
        b_m
    }.
    \label{eq:transfer_error}
\end{equation}
An error of zero indicates identical predicted responses, whereas an error of one corresponds to the mean squared response distance between distinct reference images.

Averaging errors first across evaluating models in family $G$ and then across targets in $\mathcal{I}_{\kappa}^{s,\rho}$ yields the directional transfer matrix:
\begin{equation}
    \left(T^{s\to u,\rho,\kappa}\right)_{FG}
    =
    -\frac{1}{|\mathcal{I}_{\kappa}^{s,\rho}|}
    \sum_{i\in\mathcal{I}_{\kappa}^{s,\rho}}
    \frac{1}{|\mathcal{M}_G|}
    \sum_{m\in\mathcal{M}_G}
    e_{i,F,m}^{s\to u,\rho}.
    \label{eq:directional_transfer}
\end{equation}
The $L\times L$ matrix $T^{s\to u,\rho,\kappa}$ has generating families as rows and evaluating families as columns. Higher scores, approaching zero, indicate better predicted-response preservation.

\subsection{Family-Wide and Pair-Specific Decomposition}
\label{sec:decomposition}

For a fixed analysis condition $(s,u,\rho,\kappa)$, we abbreviate the transfer matrix $T^{s\to u,\rho,\kappa}$ as $T$. We separate average bidirectional transfer from directional imbalance as
\begin{equation}
    S=\frac{T+T^{\mathsf T}}{2},
    \qquad
    A=\frac{T-T^{\mathsf T}}{2}.
    \label{eq:transfer_decomposition}
\end{equation}
The symmetric component $S$ averages transfer across the two family directions, whereas the antisymmetric component $A$ captures half their signed difference. For cross-subject evaluations, transposition exchanges generating and evaluating family labels while keeping subjects $s$ and $u$ unchanged.

To distinguish family-wide tendencies from pair-specific deviations, we fit an additive model to the off-diagonal entries of $T$~\cite{hoff2021additive}:
\begin{equation}
    \begin{split}
        T_{FG}
        &=\mu+r_F+c_G+R_{FG}, \qquad F\ne G,\\
        (\hat{\mu},\hat{\mathbf r},\hat{\mathbf c})
        &=
        \operatorname*{arg\,min}_{\mu,\mathbf r,\mathbf c}
        \sum_{F\ne G}
        \left(T_{FG}-\mu-r_F-c_G\right)^2\\
        &\hspace{6mm}
        \text{subject to }
        \sum_F r_F=\sum_G c_G=0.
    \end{split}
    \label{eq:additive_effects}
\end{equation}
Here, $\mu$ is the mean off-diagonal transfer score, while $r_F$ and $c_G$ represent the general generating- and evaluating-family effects, respectively. The residual $R_{FG}$ captures pair-specific deviations from these effects. We omit hats on the fitted effects below.

We define each family's overall transfer level and relative generator--evaluator role as
\begin{equation}
    p_F=\frac{r_F+c_F}{2},
    \qquad
    q_F=\frac{r_F-c_F}{2}.
    \label{eq:level_role}
\end{equation}
For $F\ne G$, define the symmetric and antisymmetric residuals as $R_{S,FG}=(R_{FG}+R_{GF})/2$ and $R_{A,FG}=(R_{FG}-R_{GF})/2$. The transfer components then satisfy
\begin{equation}
    \begin{split}
        S_{FG}&=\mu+(p_F+p_G)+R_{S,FG},\\
        A_{FG}&=(q_F-q_G)+R_{A,FG}.
    \end{split}
    \label{eq:interpretable_components}
\end{equation}
Unlike the pairwise matrices $S$ and $A$, the effects $p$ and $q$ summarize each family's general transfer level and directional role across partners. The residual components capture the remaining pair-specific structure.

To quantify the contribution of family-wide effects to the symmetric and antisymmetric transfer components, we define their squared-magnitude fractions over unordered family pairs:
\begin{equation}
    \begin{split}
        \eta_{\mathrm{S,level}}
        &=
        \frac{\sum_{F<G}(p_F+p_G)^2}
        {\sum_{F<G}(S_{FG}-\mu)^2},\\
        \eta_{\mathrm{A,role}}
        &=
        \frac{\sum_{F<G}(q_F-q_G)^2}
        {\sum_{F<G}A_{FG}^2}.
    \end{split}
    \label{eq:energy_fractions}
\end{equation}

To summarize the antisymmetric role contribution across source-self evaluations, let $\mathcal{H}_{\mathrm{self}}$ denote the set of source-self analysis conditions. The pooled role fraction is
\begin{equation}
    \eta_{\mathrm{A,role}}^{\mathrm{pooled}}
    =
    \frac{
        \sum_{h\in\mathcal{H}_{\mathrm{self}}}
        \sum_{F<G}(q_F^{(h)}-q_G^{(h)})^2
    }{
        \sum_{h\in\mathcal{H}_{\mathrm{self}}}
        \sum_{F<G}(A_{FG}^{(h)})^2
    }.
    \label{eq:pooled_role_fraction}
\end{equation}
These fractions quantify the relative squared magnitude of family-wide components, rather than agreement between family groupings.

\section{Experiments and Results}
\label{sec:experiments_and_results}

\subsection{Experimental Setup}
\label{sec:experimental_setup}

\paragraph{Data and regions of interest.}
We used preprocessed single-trial fMRI responses from all eight NSD subjects~\cite{allen2022a} (beta version 3; subject-native functional space at 1.8-mm isotropic resolution). Bilateral V1 and hV4 were defined using NSD's retinotopic visual-area labels (877--1,350 and 397--687 voxels across subjects, respectively). Images were split by identity into training, validation, and test sets in an 8:1:1 ratio, with identical splits across ROIs and independent splits across subjects.

\paragraph{Encoding-model configuration.}
We used 80 pretrained image-classification backbones from \texttt{torchvision}, grouped by architectural lineage into 15 families: AlexNet, VGG, Inception, SqueezeNet, ResNet, DenseNet, EfficientNet, RegNet, ConvNeXt, MobileNet, MNASNet, ShuffleNet, ViT, Swin, and MaxViT. Each family contained 1--15 backbone variants (Supplementary Table~\ref{tab:model_families}). Candidate-layer activations were flattened across all non-batch dimensions, including all ViT tokens, and reduced to 256 principal components. Voxel-wise ridge regularization was selected from $\{10^1,\ldots,10^{10}\}$ by five-fold cross-validation, and layers were selected by mean validation-set Pearson correlation across valid voxels. Fitting and refitting followed Sec.~\ref{sec:encoding_model_fitting}, yielding 1,280 encoding models.

\paragraph{Target allocation and metamer synthesis.}
Metamers were generated using models fitted to subjects 01, 02, and 05. Targets were selected from each generating subject's test set using the family-balanced prediction-agreement score (Sec.~\ref{sec:metamer_generation}). For each ROI, the top/bottom target counts were 100/50 for subject 01 and 50/25 each for subjects 02 and 05. One metamer per family for each of the 600 generating-subject--ROI--target combinations yielded 9,000 images. We analyzed 800-iteration checkpoints from Adam optimization in $3\times224\times224$ pixel space (learning rate $0.01$, $\beta_1=0.9$, $\beta_2=0.999$, $\epsilon=10^{-8}$). Pixels were initialized from $\mathcal{N}(0.5,0.05^2)$ and clamped to $[0,1]$, with a shared initialization across families for each generating-subject--ROI--target combination.

\paragraph{Transfer evaluation and statistical analysis.}
Each metamer was evaluated using models fitted to all eight subjects within the same ROI, yielding 96 directional $15\times15$ transfer matrices (12 source-self and 84 cross-subject conditions). Target-block bootstrap assessed target-sampling uncertainty and diagonal specificity, while family-label permutations tested cross-subject correspondence. Holm correction was applied separately to the 96 specificity tests and 96 correspondence tests. Transfer-profile reliability was assessed using repeated target splits with Spearman--Brown correction. Resampling details and controls excluding targets included in an evaluating subject's final-fitting data are reported in Supplementary Sec.~\ref{sec:supp_transfer_reliability}.

\subsection{Encoding-Model Performance and Metamer Fidelity}
\label{sec:metamer_results}

We assessed the held-out predictive performance of the 1,280 fitted encoding models. For each subject and ROI, the 80 model-level mean voxel-wise Pearson correlations were summarized by their median. Across subjects, these medians ranged from 0.291 to 0.476 in V1 and from 0.239 to 0.424 in hV4. The corresponding noise-ceiling-normalized rectified squared correlations ranged from 0.478 to 0.588 in V1 and from 0.318 to 0.623 in hV4.

We assessed source-response matching using the family-averaged synthesis loss (Eq.~\ref{eq:family_objective}). At 800 iterations, condition-wise median losses across target--family units ranged from $3.69\times10^{-5}$ to $1.75\times10^{-4}$, with corresponding 95th percentiles ranging from $1.30\times10^{-4}$ to $1.76\times10^{-3}$ across the 12 generating-subject--ROI--cohort conditions. These results indicate low aggregate response-matching errors for most metamers, although exact preservation was not guaranteed for every constituent model.

\begin{figure}[t]
    \centering
    \includegraphics[width=\columnwidth]{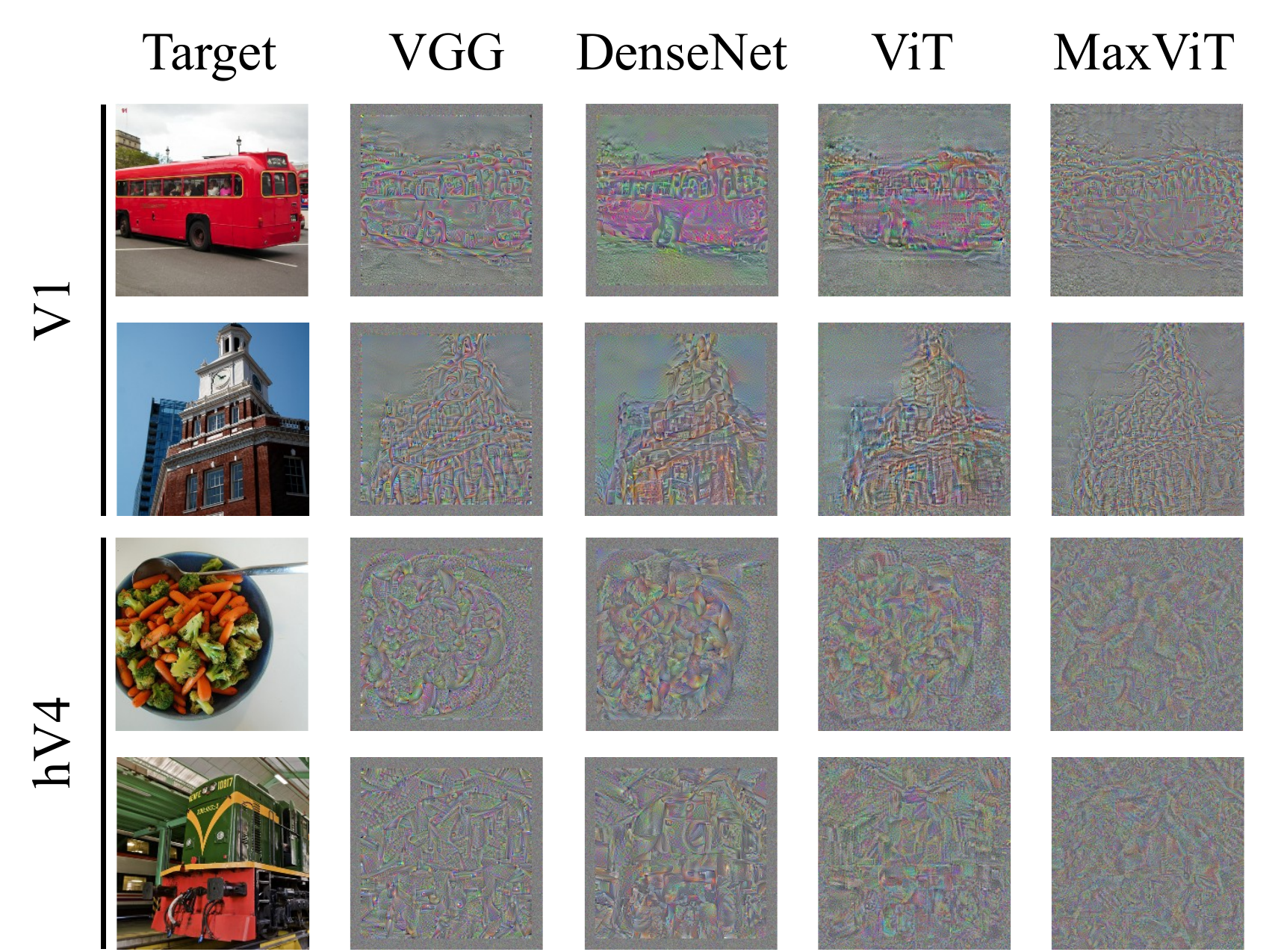}
    \caption{\textbf{Examples of family-constrained encoding-model metamers.}
    Two target images selected from the top-ranked cohort for each ROI and corresponding metamers generated using the VGG, DenseNet, ViT, and MaxViT families. All encoding models were fitted to subject 01. For each target, the generating families shared the same initialization but imposed different model-specific response-preservation constraints.}
    \label{fig:metamer_examples}
\end{figure}

Figure~\ref{fig:metamer_examples} illustrates visually distinct family-dependent metamers for the same target images. Detailed encoding-model prediction statistics and synthesis-loss summaries are reported in Supplementary Sec.~\ref{sec:supp_encoding_fidelity}.

\subsection{Family-Specific and Reproducible Metamer Transfer}
\label{sec:directional_results}

\begin{figure*}[t]
    \centering
    \includegraphics[width=\textwidth]{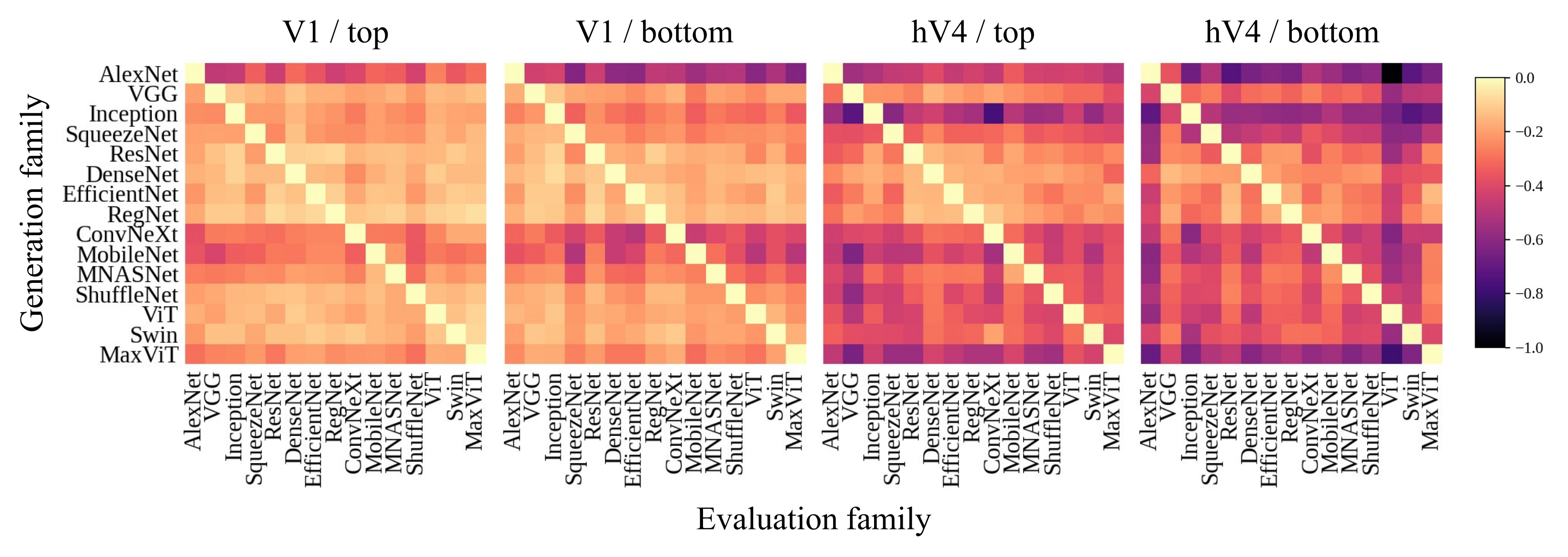}
    \caption{\textbf{Family-specific directional metamer transfer.}
    Representative $15\times15$ source-self transfer matrices for subject 01 in V1 and hV4, including top- and bottom-ranked target cohorts. Rows and columns denote generating and evaluating families, respectively. All matrices share the same family order and color scale; higher scores (closer to zero) indicate better predicted-response preservation.}
    \label{fig:transfer_matrices}
\end{figure*}

We first examined whether family-constrained metamers preferentially preserved predicted responses within their generating families (Fig.~\ref{fig:transfer_matrices}). Across all 96 conditions, mean diagonal and off-diagonal transfer scores were $-0.087$ and $-0.413$, respectively, yielding a mean advantage of $0.327$. This advantage was positive and Holm-significant in every condition. Although expected from the synthesis constraints, the diagonal advantage motivated examination of reproducible structure among the off-diagonal entries.

Across repeated target splits, the mean Spearman--Brown-corrected split-half reliability of the 210 ordered off-diagonal entries was $0.977$ across the 96 conditions (range: $0.898$--$0.998$), indicating stable cross-family transfer profiles across target subsets.

We further tested correspondence across evaluating subjects while holding the generating subject, target identities, and metamers fixed. Across 84 comparisons, the mean Pearson correlation between ordered off-diagonal transfer profiles was $0.872$ (range: $0.641$--$0.967$), with significant family-label correspondence in all comparisons after Holm correction across 96 tests.

Comparisons between native source-self profiles from different generating subjects yielded a mean off-diagonal correlation of $0.740$ (range: $0.532$--$0.895$), with significant family-label correspondence in all 12 comparisons under the same correction. Because generating subjects, selected targets, and metamers changed simultaneously, these comparisons do not isolate the effect of subject identity.

Complete transfer matrices, statistical tests, reliability analyses, cross-subject comparisons, and fitting-set overlap controls are reported in Supplementary Sec.~\ref{sec:supp_transfer_reliability}.

\subsection{Family-Wide Organization of Symmetric and Antisymmetric Transfer}
\label{sec:grouping_interpretation}

\begin{figure*}[t]
    \centering
    \includegraphics[width=\textwidth]{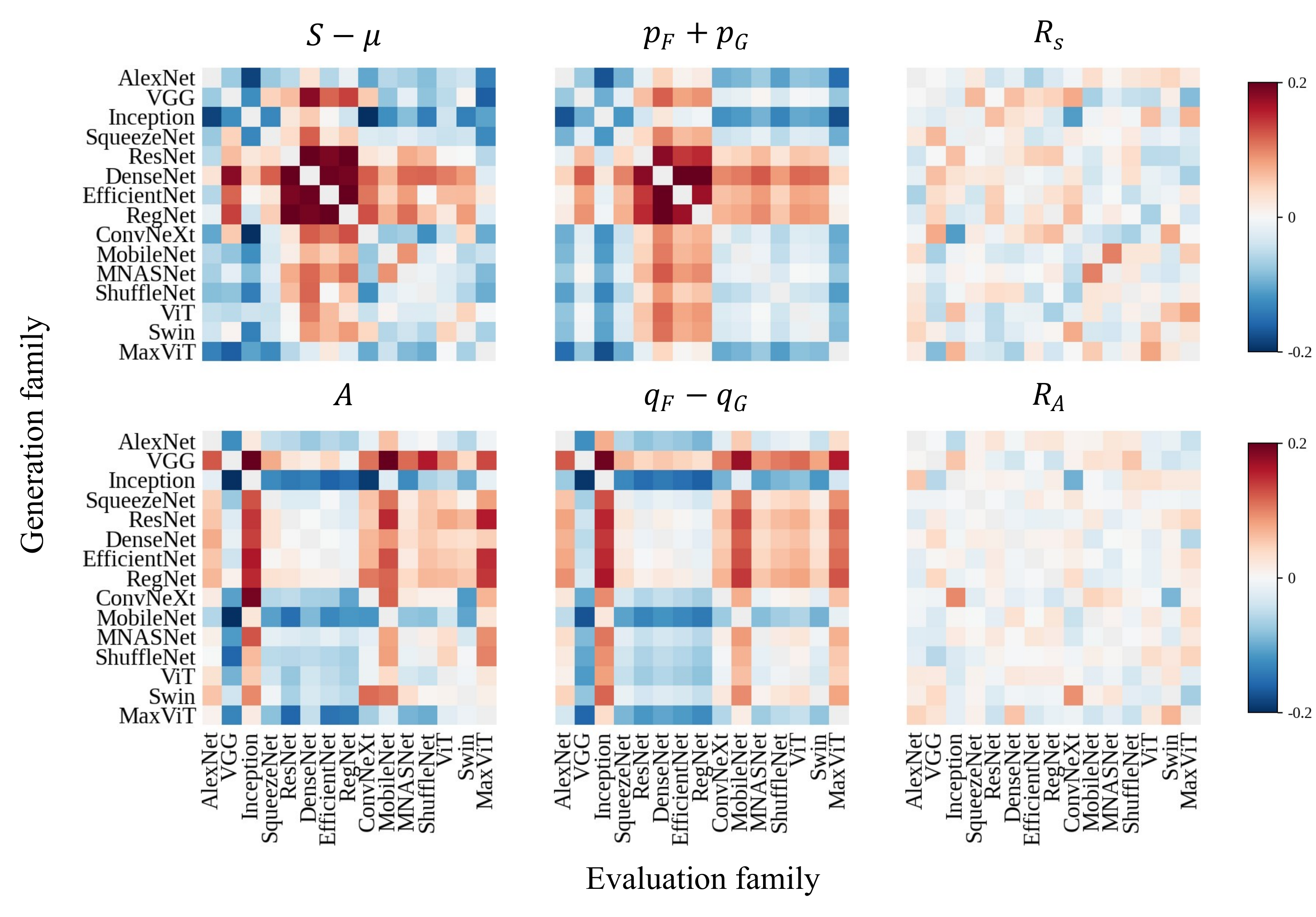}
    \caption{\textbf{Family-wide organization of symmetric and antisymmetric transfer.}
    Decomposition of the source-self transfer matrix for subject 01, hV4, and the top-ranked target cohort. The upper row shows centered symmetric transfer $S-\mu$, the family-wide level component $p_F+p_G$, and the symmetric residual $R_S$. The lower row shows antisymmetric transfer $A$, the family-wide role component $q_F-q_G$, and the antisymmetric residual $R_A$. Rows and columns denote generating and evaluating families, respectively. Diagonal entries are masked, and all six matrices use the same zero-centered color scale.}
    \label{fig:family_wide_decomposition}
\end{figure*}

We next examined whether the reproducible transfer structure could be characterized by family-wide effects. Across the 96 transfer conditions, the mean condition-wise median Spearman--Brown-corrected split-half reliabilities were $0.981$ for the symmetric component $S$ and $0.972$ for the antisymmetric component $A$, indicating that both bidirectional transfer and directional imbalance were stable across target subsets.

Using the additive decomposition in Sec.~\ref{sec:decomposition}, we separated $S$ and $A$ into family-wide level and role components and their pair-specific residuals (Fig.~\ref{fig:family_wide_decomposition}). In the representative subject 01/hV4/top condition, the level component accounted for $81.9\%$ of the centered symmetric squared magnitude. Across the 12 source-self conditions, the mean condition-wise level fraction was $82.8\%$. For the antisymmetric component, family-wide role differences $q_F-q_G$ accounted for $90.7\%$ of the pooled squared magnitude, with a target-block bootstrap 95\% CI of $[89.91\%,91.31\%]$. These fractions use different aggregation schemes and should not be directly compared. Their large values are empirical properties of the transfer matrices, rather than a necessary consequence of the decomposition.

The family-wide effects were also reproducible. Across the 12 source-self conditions, mean condition-wise median Spearman--Brown-corrected split-half reliabilities were $0.987$ for the overall transfer level $p$ and $0.965$ for the directional role $q$. When the generating subject and metamers were held fixed and only the evaluating subject changed, profile similarities averaged $0.911$ for $p$ and $0.880$ for $q$.

In subject 01/hV4/top, VGG and MobileNet had similar transfer levels but opposite roles. VGG-generated metamers preserved target responses relatively well across other families, whereas VGG’s responses were less well preserved by other families’ metamers; MobileNet showed the converse pattern. Thus, generating broadly transferable metamers and preserving responses to other families’ metamers are distinct properties.

Additional analyses of symmetric and antisymmetric profile dissimilarities, family clustering, level- and role-based groupings, and target-sampling sensitivity are reported in the supplementary material (Sec.~\ref{sec:supp_family_geometry}).

\subsection{Reliable Pair-Specific Residual Structure}
\label{sec:residual_results}

We next examined the magnitude and reproducibility of pair-specific transfer residuals. Across the 96 transfer conditions, the mean condition-wise root-mean-square (RMS) magnitude was $0.1174$ for the antisymmetric component $A$ and $0.0333$ for its residual $R_A$. The unweighted mean of the condition-wise RMS ratios $\operatorname{RMS}(R_A)/\operatorname{RMS}(A)$ was $0.296$ (range: $0.182$--$0.477$). Thus, the pair-specific directional component was substantially smaller than the original antisymmetric component but remained nonzero.

To assess residual reproducibility, we refitted the additive effects independently within each half of repeated target splits before computing $R_S$ and $R_A$. Across the 96 conditions, the mean condition-wise median Spearman--Brown-corrected split-half reliabilities were $0.973$ for $R_S$ and $0.960$ for $R_A$. Split-half similarity was measured by Pearson correlation for $R_S$ and signed Frobenius cosine similarity for $R_A$. These results indicate that residual transfer profiles retained reproducible structure beyond family-wide effects, although profile-level reliability does not establish significant differences for individual family pairs.

We further assessed cross-subject correspondence while holding the generating subject, target images, and metamers fixed. Across 84 comparisons, mean residual-profile similarities were $0.786$ for $R_S$ and $0.674$ for $R_A$, lower than those of the corresponding raw components $S$ and $A$ ($0.889$ and $0.861$). Comparisons between native source-self conditions from different generating subjects yielded lower mean residual similarities of $0.590$ and $0.461$, respectively, across 12 comparisons. Because these comparisons also changed the selected targets and generated images, the differences cannot be attributed solely to subject identity.

Thus, pair-specific residuals were highly reproducible across target subsets but less consistently preserved across evaluating subjects, particularly for directionality. We interpret them as reproducible family-pair-specific relationships rather than evidence for a universal architecture taxonomy.

Condition-wise residual statistics, residual clustering analyses, and robustness checks using alternative transfer metrics, evaluator-model perturbations, target subsampling, and initialization controls are reported in the supplementary material (Sec.~\ref{sec:supp_residual_robustness}).

\subsection{Predictive Value on Held-Out Targets}
\label{sec:heldout_prediction}

\begin{table}[t]
    \centering
    \caption{\textbf{Predictive value of pair-specific residuals on held-out targets.}
    Mean MSE from five-fold target cross-validation across 12 source-self conditions.
    Reductions compare family-wide predictors with their residual-enhanced counterparts
    ($M_2\to M_4$ and $A_1\to A_2$) and were positive in all 12 conditions.
    Lower MSE indicates better prediction.}
    \label{tab:heldout_prediction}
    \small
    \setlength{\tabcolsep}{3pt}
    \begin{tabular}{@{}lccc@{}}
        \hline
        Task & Family-wide & + Residuals & Reduction \\
        \hline
        Full & 0.06007 & 0.05623 & 6.4\% \\
        Antisym. & 0.01895 & 0.01763 & 7.0\% \\
        \hline
    \end{tabular}
\end{table}

We examined whether family-wide effects and pair-specific residuals could predict transfer scores for target images not used to estimate these components. Five-fold target cross-validation was performed separately within each of the 12 source-self conditions, with all decomposition effects fitted using training targets only. For full transfer, we compared a global-mean baseline ($M_0$), a family-wide predictor ($M_2$), and a predictor incorporating both symmetric and antisymmetric pair-specific residuals ($M_4$). For antisymmetric transfer, we compared a zero baseline ($A_0$), a family-wide role predictor ($A_1$), and its residual-enhanced counterpart ($A_2$). Held-out MSE was averaged across family pairs and targets within each condition, then equally across the 12 conditions.

For full transfer prediction, mean MSE decreased from $0.08523$ under $M_0$ to $0.06007$ under $M_2$ (Table~\ref{tab:heldout_prediction}). Including both pair-specific residual components further reduced MSE to $0.05623$ ($M_4$), a $6.4\%$ reduction relative to $M_2$. Overall, $M_4$ reduced MSE by $34.0\%$ relative to $M_0$.

For antisymmetric transfer, mean MSE decreased from $0.03193$ under $A_0$ to $0.01895$ under $A_1$, and further to $0.01763$ under $A_2$. The directional residual reduced MSE by an additional $7.0\%$ relative to $A_1$, corresponding to a $44.8\%$ reduction from $A_0$. Both residual-enhanced comparisons showed improvements in all 12 source-self conditions. Bonferroni-adjusted simultaneous 95\% confidence intervals for these paired reductions excluded zero.

These results indicate that family-wide effects capture transfer structure that generalizes across selected target images, while pair-specific residuals provide additional predictive information beyond in-sample matrix reconstruction. This generalization is limited to held-out targets within the selected cohorts and the same encoding-model families.

Detailed predictor definitions, intermediate model results, cross-validation procedures, and bootstrap confidence intervals are reported in the supplementary material (Sec.~\ref{sec:supp_heldout_prediction}).

\section{Conclusion}
\label{sec:conclusion}

\paragraph{Summary and discussion.}
We used family-constrained metamers to characterize directional response-preserving transfer among encoding-model families fitted to human V1 and hV4. Transfer profiles were reproducible across target subsets and largely preserved when the same metamers were evaluated using models fitted to different subjects. Family-wide transfer levels and generator--evaluator roles captured much of the symmetric and antisymmetric structure, respectively, while pair-specific residuals remained reliable and improved prediction on held-out targets. These findings identify stable transfer relationships beyond family-wide effects, rather than merely in-sample variation. They characterize the complete fitted image-to-response mappings, including pretrained features, dimensionality reduction, and voxel-wise readouts, without isolating architectural mechanisms.

\paragraph{Limitations and future work.}

Results are conditional on three generating subjects, prediction-agreement-selected cohorts, and the tested encoding-model families. Families contain 1–15 models, and a fixed 800-iteration budget does not guarantee comparable source-response matching. Thus, directional roles may partly reflect unequal synthesis constraints or optimization difficulty. Optimization from noise without natural-image priors yields non-naturalistic stimuli that may probe out-of-distribution model behavior; whether the main findings persist for more naturalistic metamers remains untested. These relationships concern fitted model predictions under the chosen metric, not equivalence in measured neural responses or human perception. Future work should test shared targets, matched-size ensembles, naturalistic synthesis, dissociate pretrained features from fitted readouts, and validate through neural or behavioral measurements.
\section{Acknowledgements and Disclosure}

This work was supported by JST Moonshot R\&D Program, Grant Number JPMJMS259G, JST SPRING, Grant Number JPMJSP2108. 

OpenAI's ChatGPT was used for editing and coding assistance. All scientific content, analyses, and conclusions were verified by the authors.
{
    \small
    \bibliographystyle{ieeenat_fullname}
    \bibliography{main}
}
\clearpage
\setcounter{page}{1}
\maketitlesupplementary
\appendix

\section{Encoding Prediction and Metamer Fidelity}
\label{sec:supp_encoding_fidelity}

We used 80 pretrained image-classification backbones from torchvision, grouped into 15 architecture families (Table~\ref{tab:model_families}), to construct subject- and ROI-specific encoding models.

\begin{table*}[t]
    \centering
    \caption{\textbf{Encoding-model families used in the experiments.}
    The 80 pretrained torchvision backbones are grouped into 15 architecture families.
    Model names correspond to their torchvision builder identifiers.}
    \label{tab:model_families}
    \small
    \setlength{\tabcolsep}{5pt}
    \renewcommand{\arraystretch}{1.08}
    \begin{tabular}{l c p{0.72\textwidth}}
        \toprule
        \textbf{Family} & \textbf{\# models} & \textbf{Constituent model variants} \\
        \midrule
        AlexNet & 1 &
        alexnet \\

        VGG & 8 &
        vgg11, vgg11\_bn, vgg13, vgg13\_bn,
        vgg16, vgg16\_bn, vgg19, vgg19\_bn \\

        Inception & 2 &
        googlenet, inception\_v3 \\

        SqueezeNet & 2 &
        squeezenet1\_0, squeezenet1\_1 \\

        ResNet & 10 &
        resnet18, resnet34, resnet50, resnet101, resnet152,
        resnext50\_32x4d, resnext101\_32x8d,
        resnext101\_64x4d, wide\_resnet50\_2,
        wide\_resnet101\_2 \\

        DenseNet & 4 &
        densenet121, densenet161, densenet169, densenet201 \\

        EfficientNet & 11 &
        efficientnet\_b0, efficientnet\_b1,
        efficientnet\_b2, efficientnet\_b3,
        efficientnet\_b4, efficientnet\_b5,
        \mbox{efficientnet\_b6}, efficientnet\_b7,
        efficientnet\_v2\_s, efficientnet\_v2\_m,
        efficientnet\_v2\_l \\

        RegNet & 15 &
        regnet\_x\_400mf, regnet\_x\_800mf,
        regnet\_x\_1\_6gf, regnet\_x\_3\_2gf,
        regnet\_x\_8gf, regnet\_x\_16gf,
        regnet\_x\_32gf,
        regnet\_y\_400mf, regnet\_y\_800mf,
        regnet\_y\_1\_6gf, regnet\_y\_3\_2gf,
        regnet\_y\_8gf, regnet\_y\_16gf,
        regnet\_y\_32gf, regnet\_y\_128gf \\

        ConvNeXt & 4 &
        convnext\_tiny, convnext\_small,
        convnext\_base, convnext\_large \\

        MobileNet & 3 &
        mobilenet\_v2, mobilenet\_v3\_small,
        mobilenet\_v3\_large \\

        MNASNet & 4 &
        mnasnet0\_5, mnasnet0\_75,
        mnasnet1\_0, mnasnet1\_3 \\

        ShuffleNet & 4 &
        shufflenet\_v2\_x0\_5, shufflenet\_v2\_x1\_0,
        shufflenet\_v2\_x1\_5, shufflenet\_v2\_x2\_0 \\

        ViT & 5 &
        vit\_b\_16, vit\_b\_32, vit\_l\_16,
        vit\_l\_32, vit\_h\_14 \\

        Swin & 6 &
        swin\_t, swin\_s, swin\_b,
        swin\_v2\_t, swin\_v2\_s, swin\_v2\_b \\

        MaxViT & 1 &
        maxvit\_t \\

        \midrule
        \textbf{Total} & \textbf{80} & \textbf{15 architecture families} \\
        \bottomrule
    \end{tabular}
\end{table*}

\subsection{Held-out encoding-model prediction}

The present study analyzed previously collected data from the Natural Scenes Dataset (NSD). The original NSD study was approved by the University of Minnesota Institutional Review Board, and written informed consent was obtained from all participants.

We evaluated held-out prediction performance for all 1,280 final-refit encoding models, comprising 80 pretrained backbones fitted separately for each of eight subjects and two ROIs. For voxel $v$, let $\rho_v$ denote the Pearson correlation between measured and predicted responses across test images, $\mathrm{NCSNR}_v$ the voxel-specific noise-ceiling signal-to-noise ratio, $N_{\mathrm{test}}$ the number of distinct test images, and $n_i$ the number of presentations contributing to the averaged response for test image $i$. The noise-ceiling-normalized rectified squared correlation was computed as
\begin{equation}
    \begin{aligned}
        s_v &= \frac{\max(\rho_v,0)^2}{\mathrm{NC}_v},\\
        \mathrm{NC}_v &= \frac{\mathrm{NCSNR}_v^2}{\mathrm{NCSNR}_v^2+N_{\mathrm{test}}^{-1}\sum_{i=1}^{N_{\mathrm{test}}}n_i^{-1}}.
    \end{aligned}
    \label{eq:supp_encoding_test_score}
\end{equation}
This measure differs from conventional residual-sum-of-squares-based explained variance and was not truncated at one. Image-specific presentation counts $n_i$ were retained to account for differences in repetition counts. For each encoding model, finite voxel-level scores were averaged, and the resulting 80 model-level means were summarized by their median and interquartile range for each subject--ROI combination (Table~\ref{tab:encoding_performance}).

\begin{table*}[t]
    \centering
    \caption{\textbf{Held-out encoding-model prediction performance.}
    Entries are medians [25th, 75th percentiles] across 80 model-level mean scores for each subject and ROI. Model-level scores average finite voxel-wise quantities. The normalized measure is defined in Eq.~\ref{eq:supp_encoding_test_score}.}
    \small
    \setlength{\tabcolsep}{4pt}
    \renewcommand{\arraystretch}{1.08}
    \begin{tabular}{lcccc}
        \toprule
        & \multicolumn{2}{c}{Normalized rectified $r^2$ / noise ceiling} & \multicolumn{2}{c}{Unnormalized mean Pearson $r$} \\
        \cmidrule(lr){2-3}\cmidrule(lr){4-5}
        Subject & V1 & hV4 & V1 & hV4 \\
        \midrule
        01 & 0.520 [0.506, 0.535] & 0.425 [0.410, 0.441] & 0.476 [0.470, 0.483] & 0.340 [0.333, 0.346] \\
        02 & 0.530 [0.518, 0.541] & 0.543 [0.527, 0.564] & 0.467 [0.462, 0.472] & 0.424 [0.416, 0.430] \\
        03 & 0.573 [0.560, 0.584] & 0.623 [0.535, 0.691] & 0.405 [0.400, 0.409] & 0.292 [0.283, 0.299] \\
        04 & 0.585 [0.571, 0.595] & 0.419 [0.395, 0.442] & 0.372 [0.368, 0.375] & 0.253 [0.245, 0.259] \\
        05 & 0.485 [0.475, 0.496] & 0.386 [0.364, 0.401] & 0.410 [0.406, 0.414] & 0.337 [0.327, 0.344] \\
        06 & 0.478 [0.466, 0.485] & 0.318 [0.297, 0.332] & 0.377 [0.373, 0.379] & 0.239 [0.229, 0.244] \\
        07 & 0.502 [0.486, 0.512] & 0.398 [0.369, 0.421] & 0.359 [0.354, 0.363] & 0.272 [0.261, 0.278] \\
        08 & 0.588 [0.572, 0.601] & 0.474 [0.463, 0.491] & 0.291 [0.287, 0.292] & 0.245 [0.242, 0.249] \\
        \bottomrule
    \end{tabular}
    \label{tab:encoding_performance}
\end{table*}

\begin{figure*}[t]
    \centering
    \includegraphics[width=\textwidth]{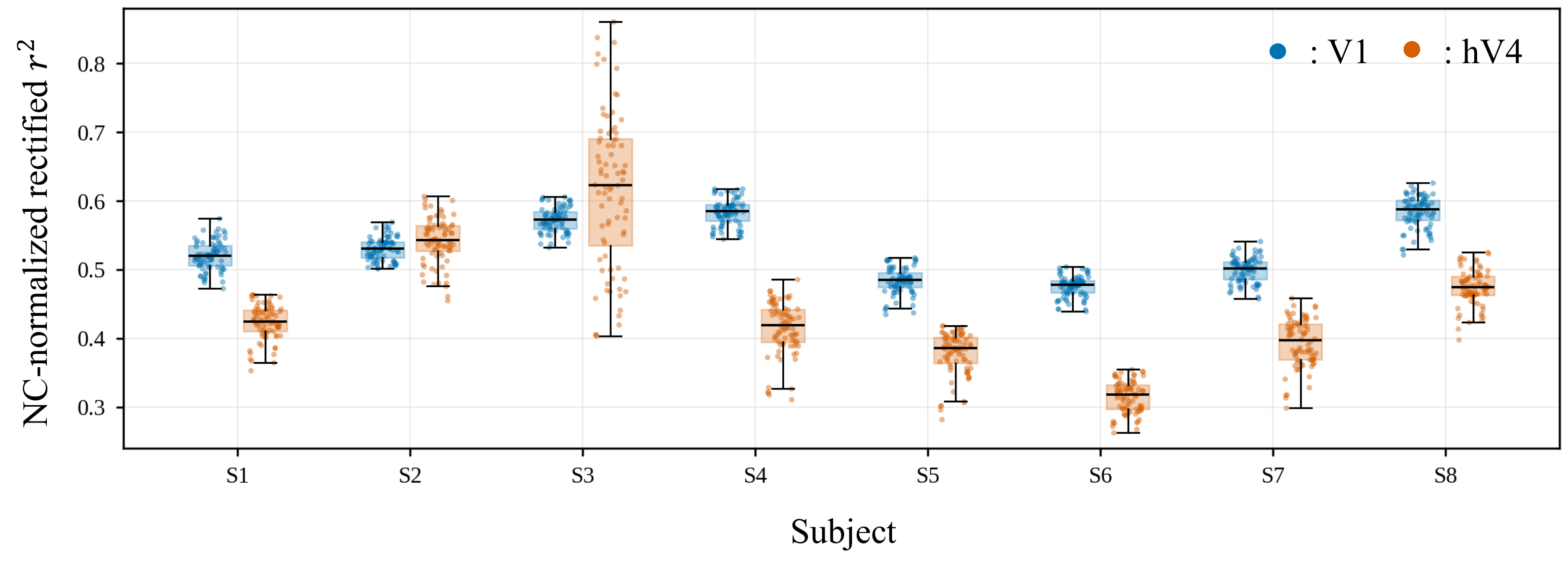}
    \caption{\textbf{Held-out encoding-model performance across subjects and ROIs.}
    Distribution of noise-ceiling-normalized rectified squared Pearson correlations for 1,280 fitted encoding models (80 models per subject and ROI). Each point represents one model-level mean across valid voxels. Medians and interquartile ranges are reported in Table~\ref{tab:encoding_performance}.}
    \label{fig:supp_encoding_performance}
\end{figure*}

Figure~\ref{fig:supp_encoding_performance} shows the distributions of model-level encoding performance. Across subjects, median unnormalized Pearson correlations ranged from 0.291 to 0.476 for V1 and from 0.239 to 0.424 for hV4. The corresponding normalized-score medians ranged from 0.478 to 0.588 and from 0.318 to 0.623, respectively. All 80 fitted models for each subject and ROI were retained for subsequent transfer evaluation.

\subsection{Source-response matching and qualitative examples}

Source-response matching was evaluated at the retained 800-iteration checkpoints using the raw-response, target-energy-normalized family synthesis objective in Eq.~\ref{eq:family_objective}. Across the 12 generating-subject--ROI--cohort conditions, median target--family-level synthesis losses ranged from $3.69\times10^{-5}$ to $1.75\times10^{-4}$. The corresponding condition-wise 95th percentiles ranged from $1.30\times10^{-4}$ to $1.76\times10^{-3}$, and maximum losses ranged from $3.64\times10^{-4}$ to $2.00\times10^{-2}$. These results indicate low synthesis losses for most target--family combinations.

The synthesis objective compares unstandardized source-model predictions normalized by target-response energy, whereas the primary transfer metric compares fit-standardized evaluating-model predictions normalized by reference-image response distance. Representative target--metamer pairs generated under different family constraints are shown in Fig.~\ref{fig:metamer_examples} of the main paper.

\section{Transfer Reliability and Cross-Subject Correspondence}
\label{sec:supp_transfer_reliability}

\subsection{Coverage and target-block inference}

The primary analysis comprised 96 transfer conditions (three generating subjects, eight evaluating subjects, two ROIs, and two target cohorts). Each condition produced a complete $15\times15$ transfer matrix, with 15 diagonal cells and 210 ordered off-diagonal cells. A total of 9,000 generated images were evaluated repeatedly with models fitted to different subjects.

To quantify uncertainty from the selected target images, bootstrap samples were drawn in target blocks: all generating families, evaluating families, and transfer directions associated with a given target were resampled together. Family-wide additive effects were refitted in each resample before computing any residual-derived quantities. Comparisons sharing target identities used matched resampling indices. Primary uncertainty summaries used 10,000 target-block bootstrap replicates unless a separate test-specific procedure is specified below. These intervals are conditional on the fitted encoding models and the evaluated subjects.

\subsection{Diagonal transfer advantage}

For condition $h$, diagonal specificity was defined as the mean diagonal transfer score minus the mean ordered off-diagonal transfer score:
\begin{equation}
    d_h = \frac{1}{L}\sum_F T_{FF}^{(h)}
    - \frac{1}{L(L-1)}\sum_{F\ne G}T_{FG}^{(h)},
\end{equation}
where $L=15$. Across 96 conditions, the mean diagonal and off-diagonal scores were $-0.086524$ and $-0.413361$, respectively, yielding a mean diagonal advantage of $0.326837$. Diagonal specificity was positive in all 96 conditions, ranging from 0.124804 to 0.816480. Across the condition-wise 95\% target-block bootstrap intervals, the smallest lower bound was 0.112920 and the largest upper bound was 0.894903.

Statistical significance was assessed using 100,000 centered studentized target-block bootstrap replicates per condition, with Holm correction across 96 tests. All 96 conditions showed significant diagonal advantages (maximum adjusted $p=9.60\times10^{-4}$), demonstrating preferential preservation of predicted responses within the generating family.

\subsection{Split-half transfer-profile reliability}

Targets were repeatedly divided into non-overlapping subsets, and each transfer matrix was recomputed separately on the two subsets. For $T$, split-half similarity was the Pearson correlation between the vectors of 210 ordered off-diagonal entries; diagonal entries were excluded. Each of 96 conditions was evaluated using 10,000 target splits. For odd-sized cohorts of 25 targets, one target was omitted from a given split to obtain equal-sized halves. A similarity $u$ was converted to its Spearman--Brown estimate $2u/(1+u)$. We summarized repeated splits by the median corrected value within each condition and then averaged these condition-wise medians with equal condition weights.

The resulting mean split-half reliability was 0.976816 across 96 conditions (condition-wise range 0.897856--0.997937); the corresponding mean before Spearman--Brown correction was 0.955871. These statistics quantify repeatability under target partitioning. The same aggregation convention is used for the component- and residual-profile reliability analyses described in Section~\ref{sec:supp_residual_robustness}.

\subsection{Cross-subject correspondence and multiplicity}

Two types of cross-subject correspondence were analyzed, both using the ordered off-diagonal transfer profiles. In the first, two native source-self matrices from different generating subjects were compared, so the subject, selected target set, and generated metamer images could all differ. In the second, the generating subject, target identities, and synthesized images were held constant, and the evaluating subject alone was changed relative to the corresponding source-self condition. The first set comprised 12 comparisons (three source-subject pairs, two ROIs, two cohorts); the second comprised 84 comparisons. Because native source-self cohorts had only about one or two overlapping canonical images per compared subject pair, matched resampling distinguished shared from nonshared targets when constructing their comparison uncertainty.

\begin{table*}[t]
    \centering
    \caption{\textbf{Off-diagonal transfer-profile correspondence.}
    Native source-self comparisons change both generating subjects and their selected targets, whereas the fixed-source comparisons reuse the same generated images and change only the evaluating subject.}
    \small
    \setlength{\tabcolsep}{4pt}
    \begin{tabular}{lrrr}
        \toprule
        Comparison & $N$ & Mean Pearson $r$ & Range \\
        \midrule
        Native source-self & 12 & 0.740 & 0.532--0.895 \\
        Same images, changed evaluator & 84 & 0.872 & 0.641--0.967 \\
        \bottomrule
    \end{tabular}
    \label{tab:supp_cross_subject_correspondence}
\end{table*}

The mean Pearson correlations were 0.740183 for native source-self comparisons (range 0.531615--0.894536) and 0.872355 for fixed-source cross-subject comparisons (range 0.640786--0.966863). Correspondence was tested with family-label permutations using 100,000 permutations per comparison, with Holm correction applied jointly across these 12+84=96 tests. All 96 comparisons were significant after correction (maximum adjusted $p=9.60\times10^{-4}$). The results support reproducible family-label organization across the tested comparisons.

\subsection{Specificity changes and fitting-set overlap}

Cross-subject evaluation did not uniformly reduce diagonal specificity. Across 84 source-self versus cross-subject comparisons, the self-minus-cross specificity difference averaged 0.068042 and ranged from $-0.280795$ to 0.513261; 25 comparisons had negative point estimates, and 13 had pointwise intervals containing zero.

Because model-fitting splits were defined independently for each subject, a generating subject's held-out target could appear in another subject's final fitting set. The audited 96 conditions contained no training-set overlaps and 12 validation-set overlap occurrences, all involving a source-target image subsequently included in final train-plus-validation refitting for certain cross-subject evaluators. For each fixed-source cross-subject comparison, the corresponding source-self and cross-subject profiles were recomputed on a matched subset excluding evaluating-model final-fitting overlaps. Mean off-diagonal correspondence remained approximately 0.872 (range approximately 0.641--0.967) across the 84 comparisons. This sensitivity analysis argues against the small observed fitting-set overlap as the sole cause of correspondence.

\subsection{Backbone-Controlled Transfer Structure and Readout Controls}
\label{sec:supp_readout_contribution}

The high cross-subject correspondence of transfer profiles raises the possibility that their shared structure primarily reflects pretrained visual backbones. To examine this possibility, we re-evaluated the same 9,000 metamers using four evaluation conditions: the original fMRI-fitted readouts, backbone-only distances in the 256-dimensional PCA feature space, Gaussian random readouts, and spectrally matched random readouts. For the backbone-only condition, we computed the mean squared Euclidean distance between 256-dimensional post-PCA feature vectors, without applying the fitted voxel-wise readout. Each distance was normalized by the corresponding mean squared distance between distinct reference images, using the same 100 reference images as in the primary transfer evaluation.

For each evaluating encoding model, let $W\in\mathbb{R}^{256\times V}$ denote the fitted readout weights mapping post-PCA features to fit-standardized voxel responses. For the Gaussian control, we independently sampled $\boldsymbol{g}_j\sim\mathcal{N}(\mathbf{0},I_{256})$ for each output voxel $j$ and defined
\begin{equation}
\widetilde{\boldsymbol{w}}_j
=
\frac{\|\boldsymbol{w}_j\|_2}{\|\boldsymbol{g}_j\|_2}
\boldsymbol{g}_j,
\end{equation}
thereby preserving the $\ell_2$ norm of each voxel's readout vector. For the spectral control, we sampled a Haar-distributed orthogonal matrix $Q\in\mathbb{R}^{256\times256}$ and set $\widetilde{W}=Q^\top W$. This transformation preserves $W^\top W$ and the singular values of $W$ while randomizing its alignment with the PCA feature axes. Random transformations were sampled independently for each evaluating-model artifact.

All conditions used the same reference-normalized transfer framework and identical target and reference image identities. The Gaussian and spectral controls recomputed their reference-normalization denominators for each random replicate. These manipulations affected only the evaluating models; the metamers remained those synthesized using the original fMRI-fitted encoding models.

To examine transfer structure not captured by a linear fit to the backbone-only transfer profile, we regressed the fMRI-fitted profile on the backbone-only profile separately for each generating-subject--ROI--cohort--evaluating-subject condition. Each regression included an intercept and used the 210 ordered off-diagonal entries of the corresponding transfer matrix. The resulting residual profile was defined as
\begin{equation}
e_{h,u}
=
t^{\mathrm{fMRI}}_{h,u}
-
\left(
\hat{\alpha}_{h,u}\mathbf{1}
+
\hat{\beta}_{h,u}t^{\mathrm{backbone}}_{h,u}
\right),
\end{equation}
where $h$ indexes the generating subject, ROI, and target cohort, $u$ denotes the evaluating subject, and $t$ represents the vectorized off-diagonal transfer profile.

For cross-subject comparisons, we excluded evaluator pairs containing the generating subject. Within each of the 12 generating-subject--ROI--cohort strata, we computed Pearson correlations between profiles for all 21 pairs among the remaining seven evaluating subjects. Correlations were averaged within each stratum and then equally across strata.

For each Gaussian and spectral replicate, we applied the same backbone regression to its randomized transfer profiles, refitting the coefficients separately for each evaluation condition. Cross-subject correlations were then computed from the resulting residual profiles using the same evaluator pairs and aggregation procedure. Thus, the observed backbone-controlled residual statistic and both random-readout null distributions were directly comparable.

The mean cross-subject correlation was 0.8619 for the original fMRI-fitted transfer profiles and 0.9225 for the backbone-only profiles (Table~\ref{tab:supp_readout_contribution}). After removing the component linearly predicted by the backbone-only profiles, the residual profiles retained substantial cross-subject correspondence (mean correlation: 0.7792). This exceeded both random-readout residual-null distributions. Neither null produced a statistic as large as the observed value in 200 random replicates (one-sided Monte Carlo $p=0.004975$ for each comparison; Holm-adjusted $p=0.009950$ across the two tests).

\begin{table}[t]
    \centering
    \caption{\textbf{Cross-subject correspondence of transfer profiles under different readout conditions.}
    Mean Pearson correlations across evaluating subjects, averaged equally over 12 generating-subject--ROI--cohort conditions. Random-readout null statistics use the same backbone-controlled residual correlations as the observed residual condition. Central 95\% intervals are empirical quantiles across 200 replicates.}
    \label{tab:supp_readout_contribution}
    \small
    \setlength{\tabcolsep}{4pt}
    \renewcommand{\arraystretch}{1.12}
    \begin{tabular}{lcc}
        \toprule
        Condition & Mean $r$ & Central 95\% interval \\
        \midrule
        fMRI-fitted & 0.8619 & -- \\
        Backbone-only & 0.9225 & -- \\
        Backbone-controlled residual & 0.7792 & -- \\
        Spectral residual null & 0.0018 & [-0.0453, 0.0667] \\
        Gaussian residual null & 0.0000 & [-0.0529, 0.0652] \\
        \bottomrule
    \end{tabular}
\end{table}

We further assessed the target-sampling reproducibility of backbone-controlled residual profiles using 10,000 balanced target splits per condition. For each split, backbone-regression coefficients estimated from one half were applied to the other half, in both directions, yielding cross-fitted residual profiles. We computed Pearson correlations across the 210 ordered off-diagonal entries for the full residual profiles and their symmetric and antisymmetric components. The reported values are uncorrected split-half correlations, averaged across splits within each condition and then equally across the 96 conditions. Unlike the reliability summaries in Secs.~B.3 and D.1, no Spearman--Brown correction or median aggregation was applied. The resulting mean correlations were 0.8813 for the full residual profile, 0.8945 for its symmetric component, and 0.8644 for its antisymmetric component.

The main result was robust to using variance-whitened PCA features for the backbone-only control, obtained by dividing each PCA coordinate by the square root of its training-set explained variance (residual cross-subject correlation: 0.8236). Recomputing fitted standardized responses from the saved PCA features and readout weights yielded a global residual correlation of 0.7791719, compared with 0.7791723 using the original response caches, agreeing to five decimal places.

These results identify reproducible cross-subject transfer structure remaining after linear adjustment for the specified backbone-only transfer profiles. The high cross-subject correspondence of backbone-only profiles also shows that strongly shared transfer structure is already present in pretrained-feature distances. Because the backbone control captures a particular feature-distance profile, the residual structure should not be interpreted as isolating readout-specific effects from all possible backbone-derived relationships.

\subsection{Complete transfer matrix atlas}
\label{sec:supp_transfer_atlas}

\begin{figure*}[p]
    \centering
    \includegraphics[width=\textwidth,height=0.79\textheight,keepaspectratio]{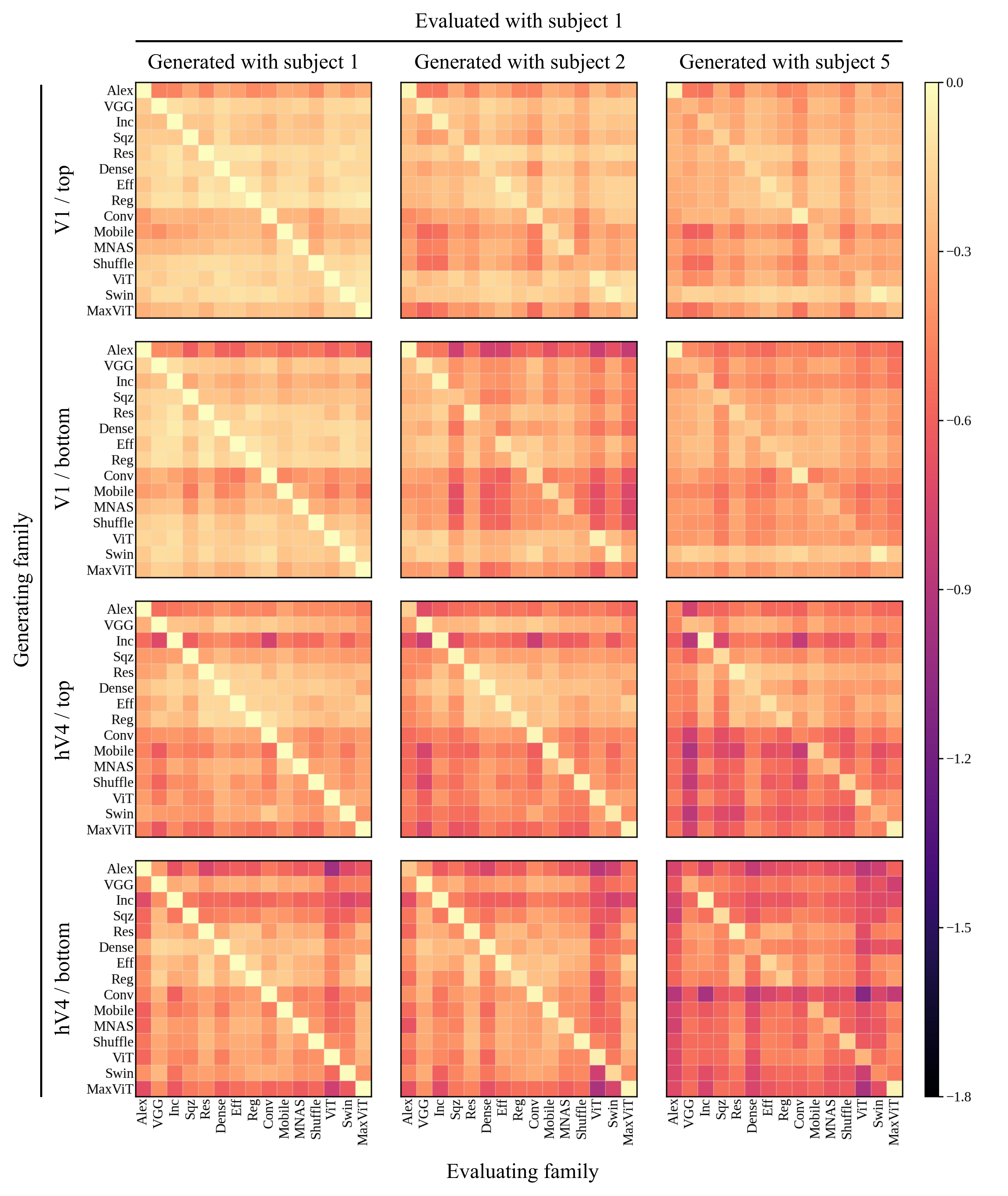}
    \caption{\textbf{Complete transfer matrices for evaluating subject 01.}
    Twelve $15\times15$ matrices are organized by generating subject (01, 02, 05), ROI (V1, hV4), and target cohort (top, bottom). Rows denote generating families and columns evaluating families; higher scores (closer to zero) indicate better response preservation. All eight atlas pages use the same family order and a common color scale. Each matrix is computed using metamers synthesized for the indicated generating subject.}
    \label{fig:supp_transfer_atlas_01}
\end{figure*}

\begin{figure*}[p]
    \centering
    \includegraphics[width=\textwidth,height=0.79\textheight,keepaspectratio]{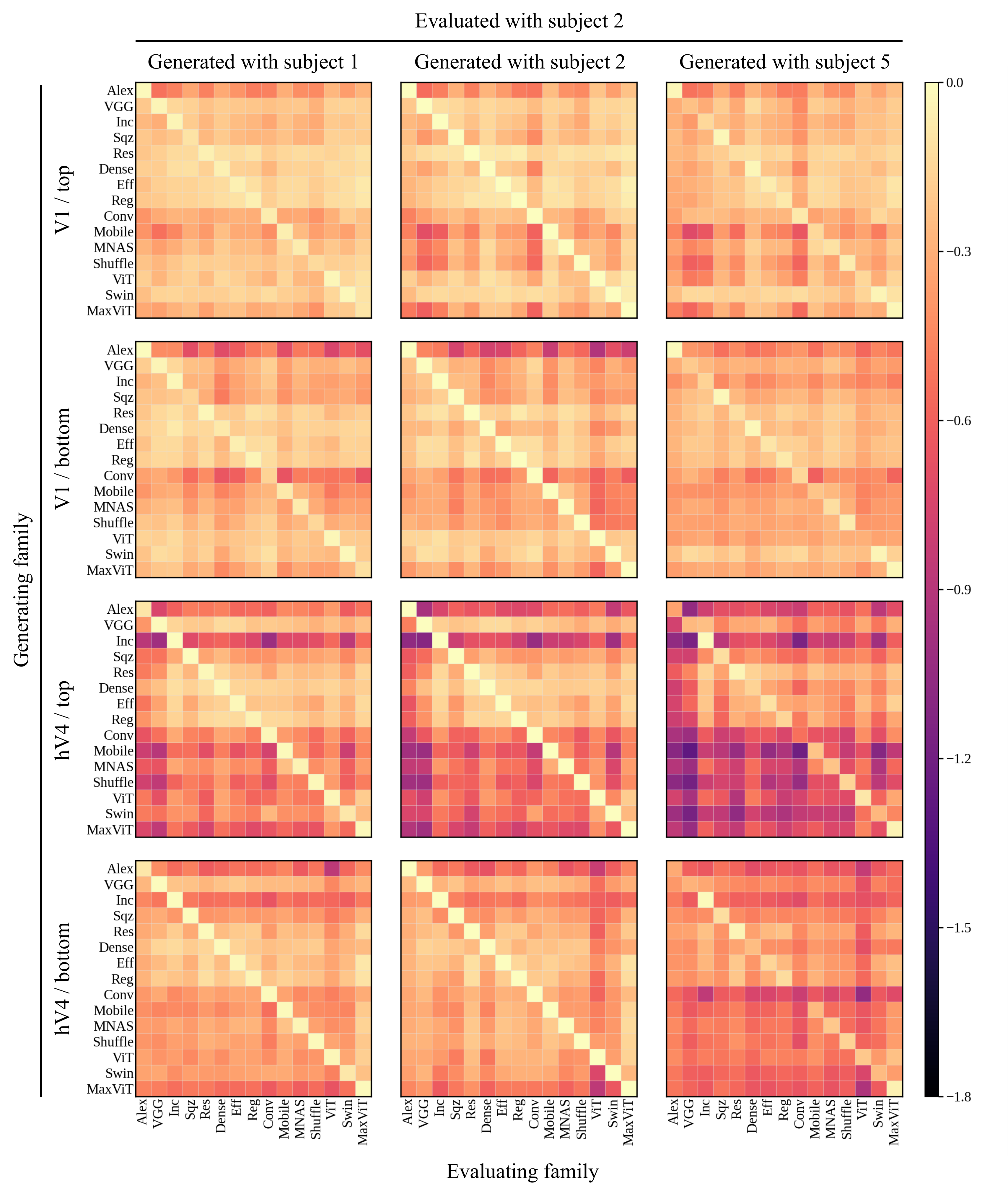}
    \caption{\textbf{Complete transfer matrices for evaluating subject 02.}
    Twelve $15\times15$ matrices are organized by generating subject (01, 02, 05), ROI (V1, hV4), and target cohort (top, bottom). Rows denote generating families and columns evaluating families; higher scores (closer to zero) indicate better response preservation. All eight atlas pages use the same family order and a common color scale. Each matrix is computed using metamers synthesized for the indicated generating subject.}
    \label{fig:supp_transfer_atlas_02}
\end{figure*}

\begin{figure*}[p]
    \centering
    \includegraphics[width=\textwidth,height=0.79\textheight,keepaspectratio]{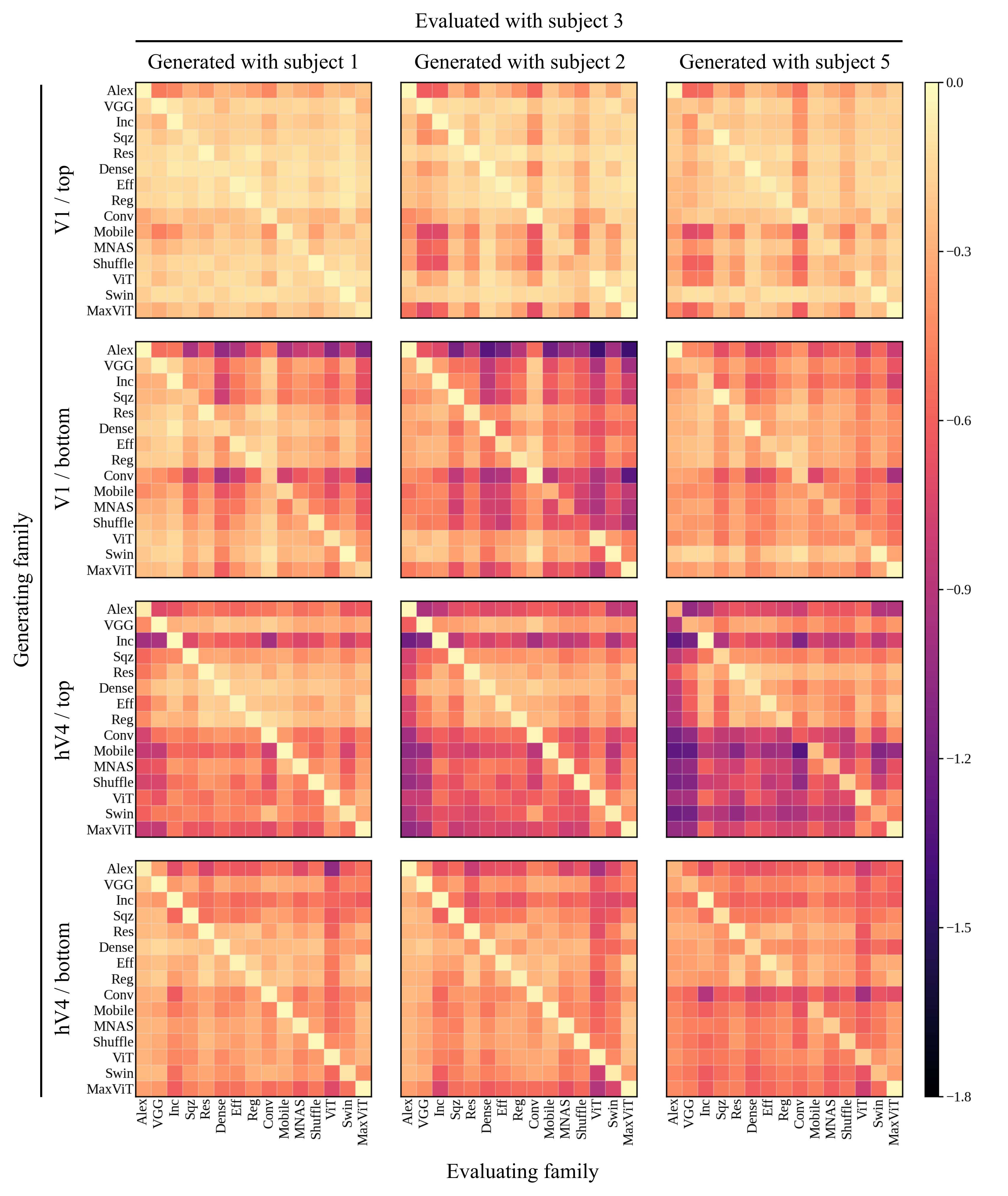}
    \caption{\textbf{Complete transfer matrices for evaluating subject 03.}
    Twelve $15\times15$ matrices are organized by generating subject (01, 02, 05), ROI (V1, hV4), and target cohort (top, bottom). Rows denote generating families and columns evaluating families; higher scores (closer to zero) indicate better response preservation. All eight atlas pages use the same family order and a common color scale. Each matrix is computed using metamers synthesized for the indicated generating subject.}
    \label{fig:supp_transfer_atlas_03}
\end{figure*}

\begin{figure*}[p]
    \centering
    \includegraphics[width=\textwidth,height=0.79\textheight,keepaspectratio]{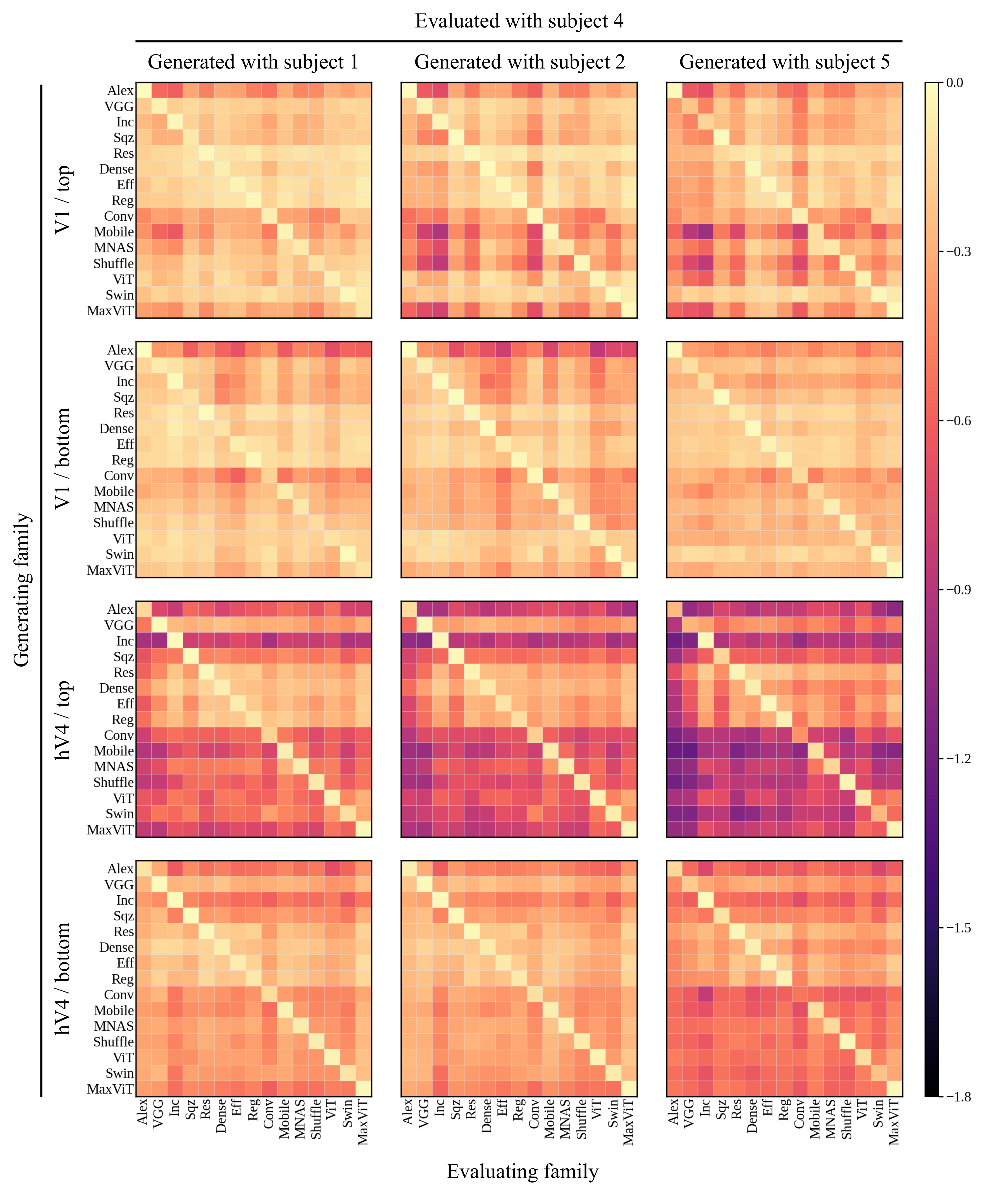}
    \caption{\textbf{Complete transfer matrices for evaluating subject 04.}
    Twelve $15\times15$ matrices are organized by generating subject (01, 02, 05), ROI (V1, hV4), and target cohort (top, bottom). Rows denote generating families and columns evaluating families; higher scores (closer to zero) indicate better response preservation. All eight atlas pages use the same family order and a common color scale. Each matrix is computed using metamers synthesized for the indicated generating subject.}
    \label{fig:supp_transfer_atlas_04}
\end{figure*}

\begin{figure*}[p]
    \centering
    \includegraphics[width=\textwidth,height=0.79\textheight,keepaspectratio]{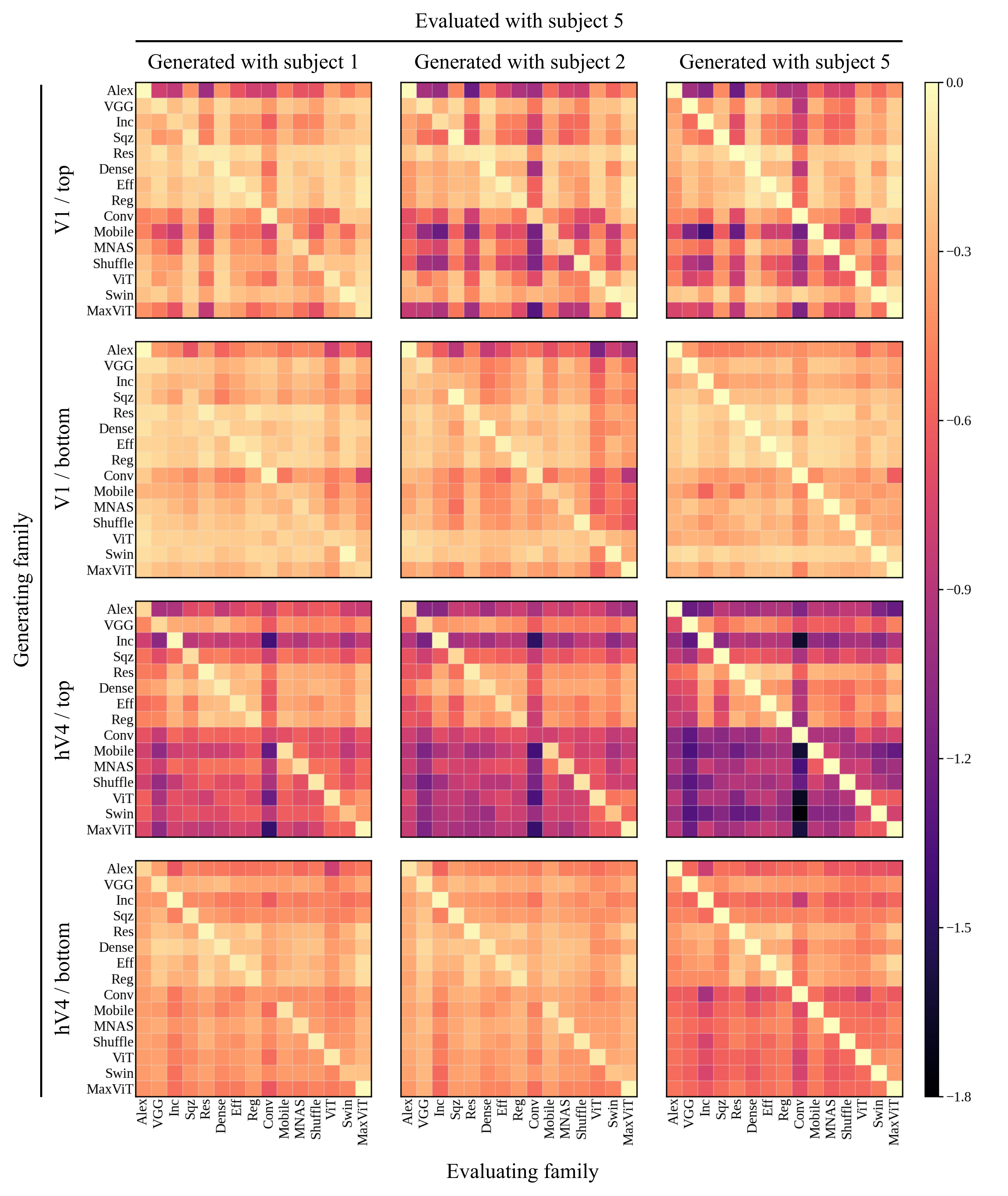}
    \caption{\textbf{Complete transfer matrices for evaluating subject 05.}
    Twelve $15\times15$ matrices are organized by generating subject (01, 02, 05), ROI (V1, hV4), and target cohort (top, bottom). Rows denote generating families and columns evaluating families; higher scores (closer to zero) indicate better response preservation. All eight atlas pages use the same family order and a common color scale. Each matrix is computed using metamers synthesized for the indicated generating subject.}
    \label{fig:supp_transfer_atlas_05}
\end{figure*}

\begin{figure*}[p]
    \centering
    \includegraphics[width=\textwidth,height=0.79\textheight,keepaspectratio]{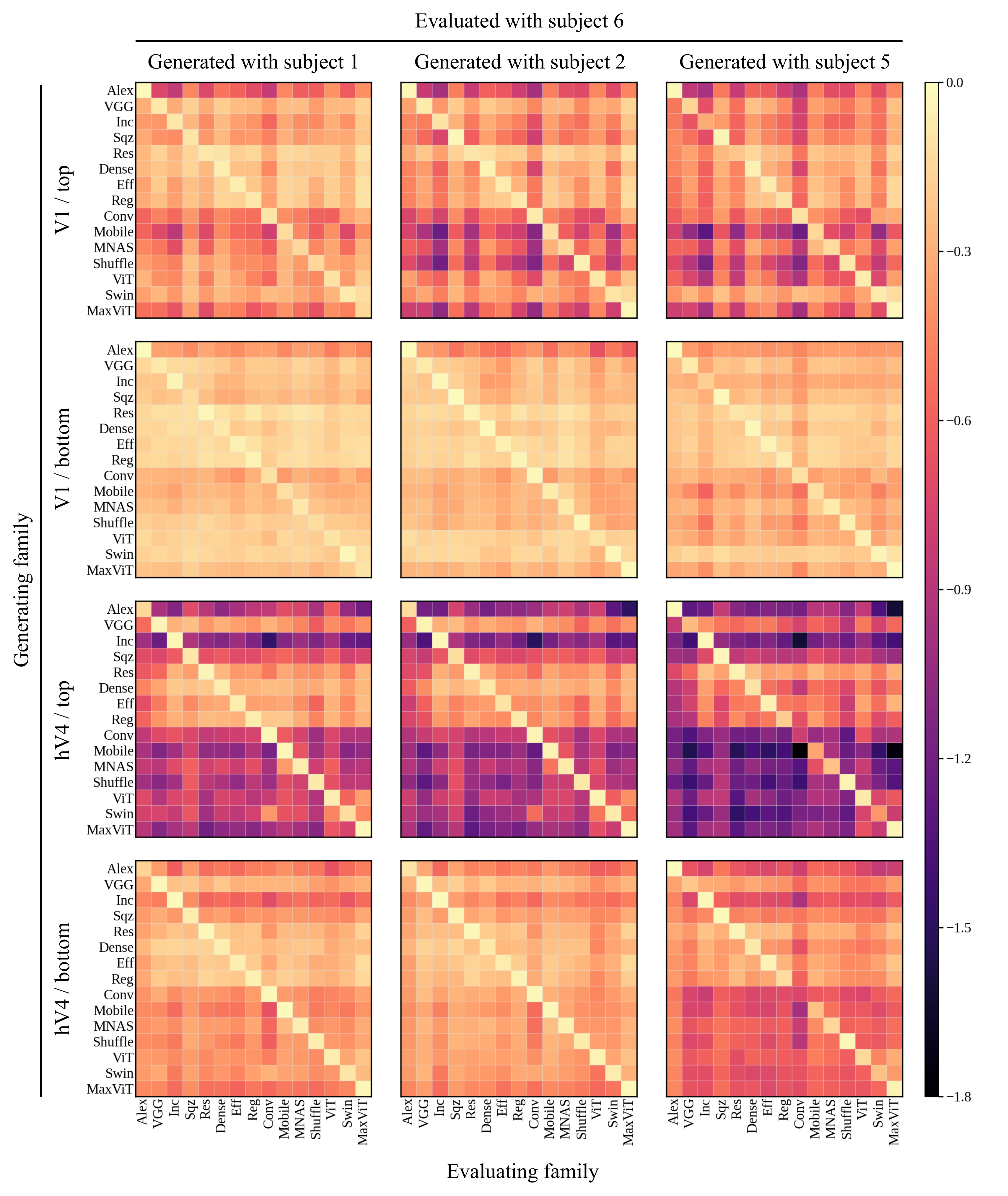}
    \caption{\textbf{Complete transfer matrices for evaluating subject 06.}
    Twelve $15\times15$ matrices are organized by generating subject (01, 02, 05), ROI (V1, hV4), and target cohort (top, bottom). Rows denote generating families and columns evaluating families; higher scores (closer to zero) indicate better response preservation. All eight atlas pages use the same family order and a common color scale. Each matrix is computed using metamers synthesized for the indicated generating subject.}
    \label{fig:supp_transfer_atlas_06}
\end{figure*}

\begin{figure*}[p]
    \centering
    \includegraphics[width=\textwidth,height=0.79\textheight,keepaspectratio]{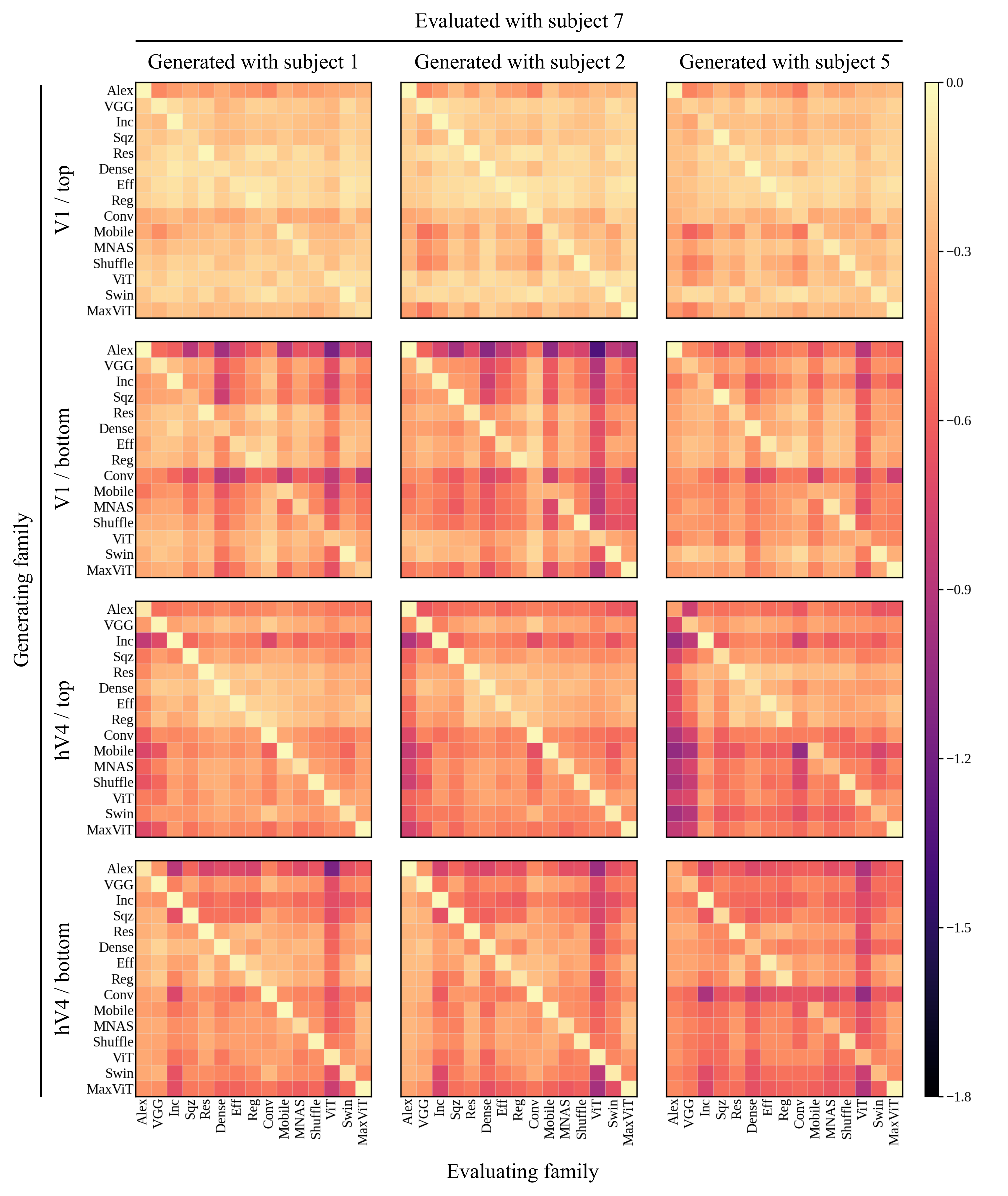}
    \caption{\textbf{Complete transfer matrices for evaluating subject 07.}
    Twelve $15\times15$ matrices are organized by generating subject (01, 02, 05), ROI (V1, hV4), and target cohort (top, bottom). Rows denote generating families and columns evaluating families; higher scores (closer to zero) indicate better response preservation. All eight atlas pages use the same family order and a common color scale. Each matrix is computed using metamers synthesized for the indicated generating subject.}
    \label{fig:supp_transfer_atlas_07}
\end{figure*}

\begin{figure*}[p]
    \centering
    \includegraphics[width=\textwidth,height=0.79\textheight,keepaspectratio]{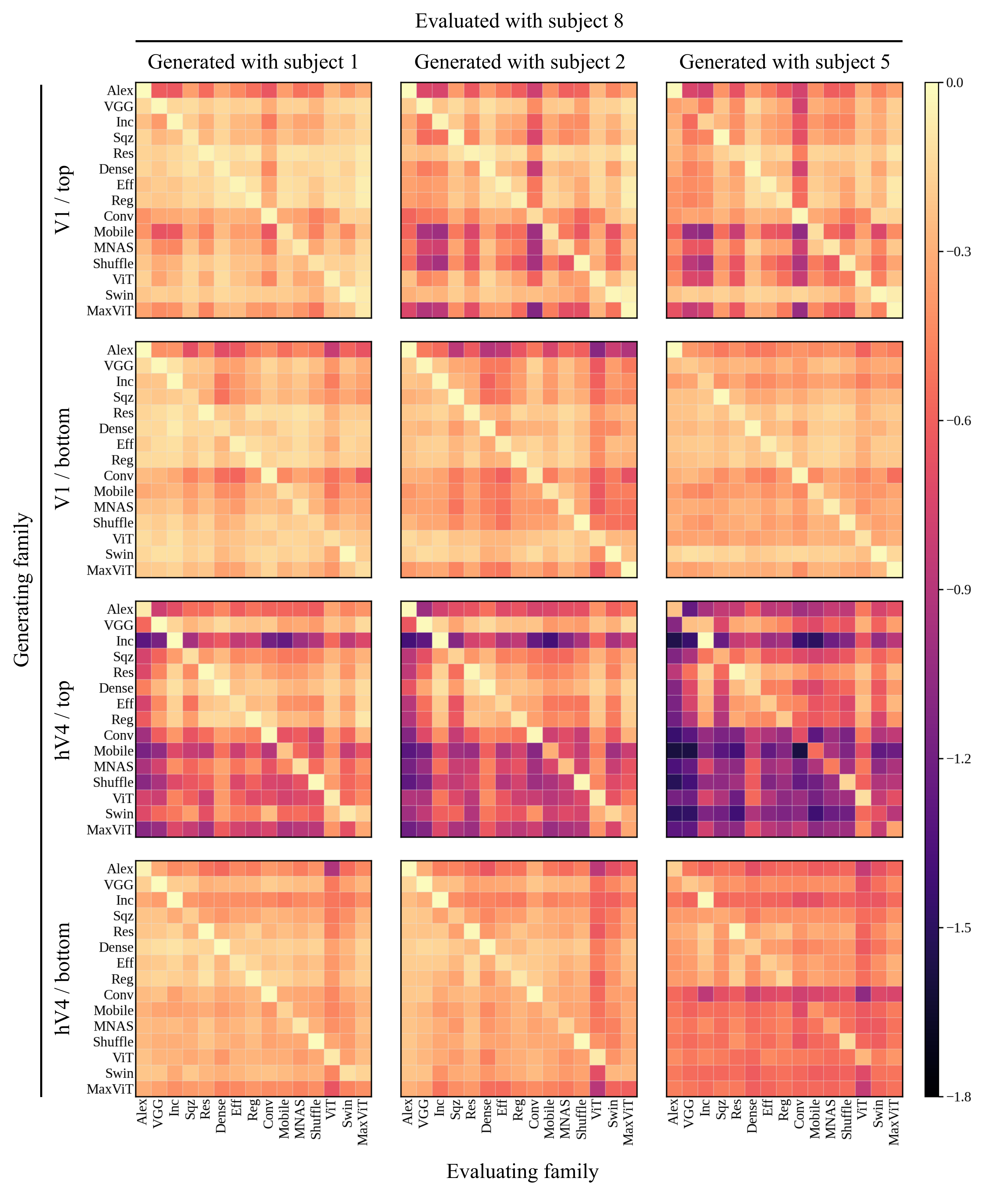}
    \caption{\textbf{Complete transfer matrices for evaluating subject 08.}
    Twelve $15\times15$ matrices are organized by generating subject (01, 02, 05), ROI (V1, hV4), and target cohort (top, bottom). Rows denote generating families and columns evaluating families; higher scores (closer to zero) indicate better response preservation. All eight atlas pages use the same family order and a common color scale. Each matrix is computed using metamers synthesized for the indicated generating subject.}
    \label{fig:supp_transfer_atlas_08}
\end{figure*}

Figures~\ref{fig:supp_transfer_atlas_01}--\ref{fig:supp_transfer_atlas_08} display all 96 directional transfer matrices, grouped by evaluating subject. Each atlas page contains 12 conditions (three generating subjects $\times$ two ROIs $\times$ two target cohorts), with the same 15 architecture families in generating rows and evaluating columns. All atlas pages use a common color scale; higher transfer scores indicate better predicted-response preservation. The panels summarize 96 evaluations of the same 9,000 synthesized images.

\clearpage

\clearpage

\clearpage

\clearpage

\clearpage

\clearpage

\clearpage

\clearpage

\section{Family-Wide Organization and Clustering}
\label{sec:supp_family_geometry}

\subsection{Additive effects and matrix-level contributions}

The off-diagonal additive decomposition in Sec.~\ref{sec:decomposition} separates family-wide generating and evaluating effects from pair-specific residuals. Under the sum-to-zero constraints, the fitted intercept $\mu$ equals the mean ordered off-diagonal entry of $T$. The least-squares residual $R$ has zero off-diagonal row and column sums, as do its symmetric and antisymmetric components $R_S$ and $R_A$. Consequently, the family-wide components $p_F+p_G$ and $q_F-q_G$ are orthogonal to their corresponding residual components under equal off-diagonal weighting. This provides an orthogonal decomposition of the squared magnitudes of the centered symmetric and antisymmetric transfer components.

Across the 12 source-self conditions, the unweighted mean of the condition-wise symmetric level fractions was 82.77\%. For the antisymmetric component, the pooled role fraction defined in Eq.~\ref{eq:pooled_role_fraction} was 0.906625, leaving a residual squared-magnitude fraction of 0.093375. In 10,000 target-block bootstrap replicates, the median pooled role fraction was 0.906106, with a 95\% percentile interval of [0.899144, 0.913084].

Both the family-wide transfer level $p$ and directional role $q$ showed high split-half reliability across the 12 source-self conditions. The averages of condition-wise median Spearman--Brown-corrected reliabilities were 0.987 for $p$ and 0.965 for $q$, with additive effects refitted independently for each split half. Fixed-source cross-subject profile similarities averaged 0.911 for $p$ and 0.880 for $q$.

In the representative subject 01/hV4/top condition, the family-wide level component accounted for approximately 81.9\% of the centered symmetric matrix's off-diagonal squared magnitude. The VGG and MobileNet families had similar level effects ($p=0.0001$ and $-0.0152$, respectively) but opposite directional roles ($q=0.0914$ and $-0.0819$). Thus, level and role capture distinct aspects of family-wide transfer organization.

\subsection{Leave-two-out profile dissimilarity}

To compare the family relationships represented by the symmetric and antisymmetric matrices, we defined the leave-two-out profile dissimilarity for $M\in\{S,A,R_S,R_A\}$ between families $F$ and $G$ as
\begin{equation}
    D_M(F,G)=\sqrt{\frac{1}{L-2}\sum_{H\in\mathcal{F}\setminus\{F,G\}}(M_{FH}-M_{GH})^2}.
    \label{eq:supp_profile_distance}
\end{equation}
This measure compares the transfer profiles of $F$ and $G$ toward the common set of third-party families, excluding their self-transfer entries and direct pairwise relationship. Smaller values indicate more similar transfer profiles.

For symmetric profiles, the additive decomposition gives $S_{FH}-S_{GH}=p_F-p_G+(R_{S,FH}-R_{S,GH})$ for every $H\notin\{F,G\}$. Because the off-diagonal row sums of $R_S$ vanish and $R_{S,FG}=R_{S,GF}$, the residual-difference sum over third-party families is zero. Expanding the squared distance therefore yields the exact identity
\begin{equation}
    D_S(F,G)^2=(p_F-p_G)^2+D_{R_S}(F,G)^2.
    \label{eq:supp_s_distance_identity}
\end{equation}
This identity decomposes symmetric-profile dissimilarity into differences in family-wide levels and pair-specific residual profiles. For antisymmetric profiles, the corresponding cross term is generally nonzero, so an analogous orthogonal distance decomposition does not follow.

We quantified the contribution of family-wide level differences to squared symmetric-profile dissimilarity within condition $h$ as
\begin{equation}
    \gamma_h=
    \frac{\sum_{F<G}(p_F^{(h)}-p_G^{(h)})^2}
    {\sum_{F<G}D_S^{(h)}(F,G)^2},
    \qquad
    \bar\gamma=
    \frac{1}{|\mathcal H_{\mathrm{self}}|}
    \sum_{h\in\mathcal H_{\mathrm{self}}}\gamma_h.
    \label{eq:supp_level_distance_fraction}
\end{equation}
Each $\gamma_h$ is computed by summing over 105 unordered family pairs before taking the ratio, and $\bar\gamma$ is the unweighted mean across source-self conditions. In the representative subject 01/hV4/top condition, $\gamma_h=0.708402$, with a target-bootstrap 95\% percentile interval of approximately [0.682, 0.731]. Across the 12 source-self conditions, $\bar\gamma=0.722$ (range: 0.597--0.847), indicating that family-wide level differences accounted for 72.2\% of the total squared symmetric-profile dissimilarity.

\subsection{Symmetric and antisymmetric family groupings}

We applied average-linkage agglomerative clustering to the full $L\times L$ leave-two-out dissimilarity matrices. The representative partition used $k=3$ clusters, with $k\in\{2,3,4,5\}$ examined for sensitivity. Agreement between partitions was quantified using the adjusted Rand index (ARI).

\begin{figure*}[t]
    \centering
    \includegraphics[width=\textwidth]{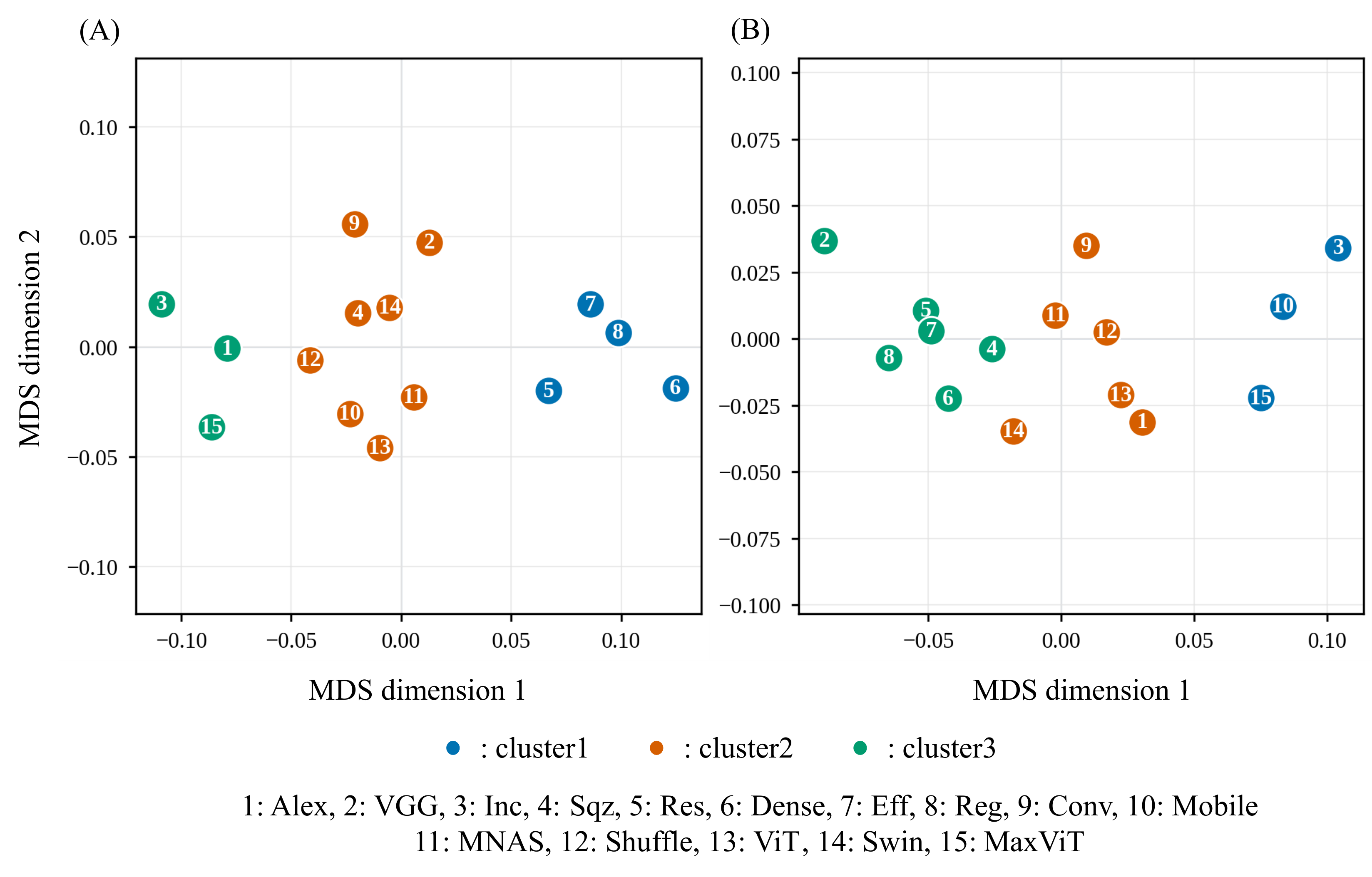}
    \caption{\textbf{Symmetric and antisymmetric family organizations.}
    Two-dimensional MDS representations of leave-two-out dissimilarities for the symmetric component $S$ and antisymmetric component $A$ in the representative subject 01/hV4/top source-self condition ($N=100$). Each point represents one architecture family, and colors indicate $k=3$ average-linkage clusters computed from the original dissimilarity matrices. Cluster colors are assigned independently in each panel. The S--A partition ARI was 0.273, and the pairwise-distance Pearson correlation was 0.498.}
    \label{fig:supp_sa_groupings}
\end{figure*}

For subject 01/hV4/top, symmetric profiles grouped ResNet, DenseNet, EfficientNet, and RegNet; VGG, SqueezeNet, ConvNeXt, MobileNet, MNASNet, ShuffleNet, ViT, and Swin; and AlexNet, Inception, and MaxViT. Antisymmetric profiles instead grouped VGG and SqueezeNet with ResNet, DenseNet, EfficientNet, and RegNet; MobileNet with Inception and MaxViT; and AlexNet, ConvNeXt, MNASNet, ShuffleNet, ViT, and Swin. The full-data S--A ARI was 0.272616, and the Pearson correlation between their pairwise dissimilarities was 0.498353 (Fig.~\ref{fig:supp_sa_groupings}). Across $k=2,3,4,5$, the S--A ARIs were $-0.124$, 0.273, 0.319, and 0.270, respectively.

\begin{table}[t]
    \centering
    \caption{\textbf{Agreement between family partitions in the subject 01 source-self conditions.}
    Adjusted Rand indices at $k=3$. $S$ and $A$ use leave-two-out profile dissimilarities, whereas $p$ and $q$ use $|p_F-p_G|$ and $|q_F-q_G|$, respectively.}
    \label{tab:sub01_grouping_agreement}
    \small
    \setlength{\tabcolsep}{4pt}
    \begin{tabular}{lrrr}
        \toprule
        Condition & $S$--$A$ & $S$--$p$ & $A$--$q$ \\
        \midrule
        V1/top & 0.716 & 0.745 & 0.769 \\
        V1/bottom & 0.546 & 1.000 & 1.000 \\
        hV4/top & 0.273 & 1.000 & 0.759 \\
        hV4/bottom & 0.121 & 1.000 & 0.638 \\
        \bottomrule
    \end{tabular}
\end{table}

The agreement between symmetric and antisymmetric partitions varied across ROI and target cohort, with S--A ARIs ranging from 0.121 to 0.716 for subject 01 (Table~\ref{tab:sub01_grouping_agreement}). In the representative hV4/top condition, the median full-versus-bootstrap partition ARI was 1.000 for $S$ (2.5--97.5 percentile range: [0.786, 1.000]) and 0.596 for $A$ ([0.358, 1.000]). Across 500 target-block bootstrap replicates, the S--A partition agreement had a median ARI of approximately 0.171 and a 2.5--97.5 percentile range of approximately [$-0.008$, 0.426].

\subsection{Profile distances, family-wide effects, and supplementary comparisons}

\begin{figure*}[t]
    \centering
    \includegraphics[width=\textwidth]{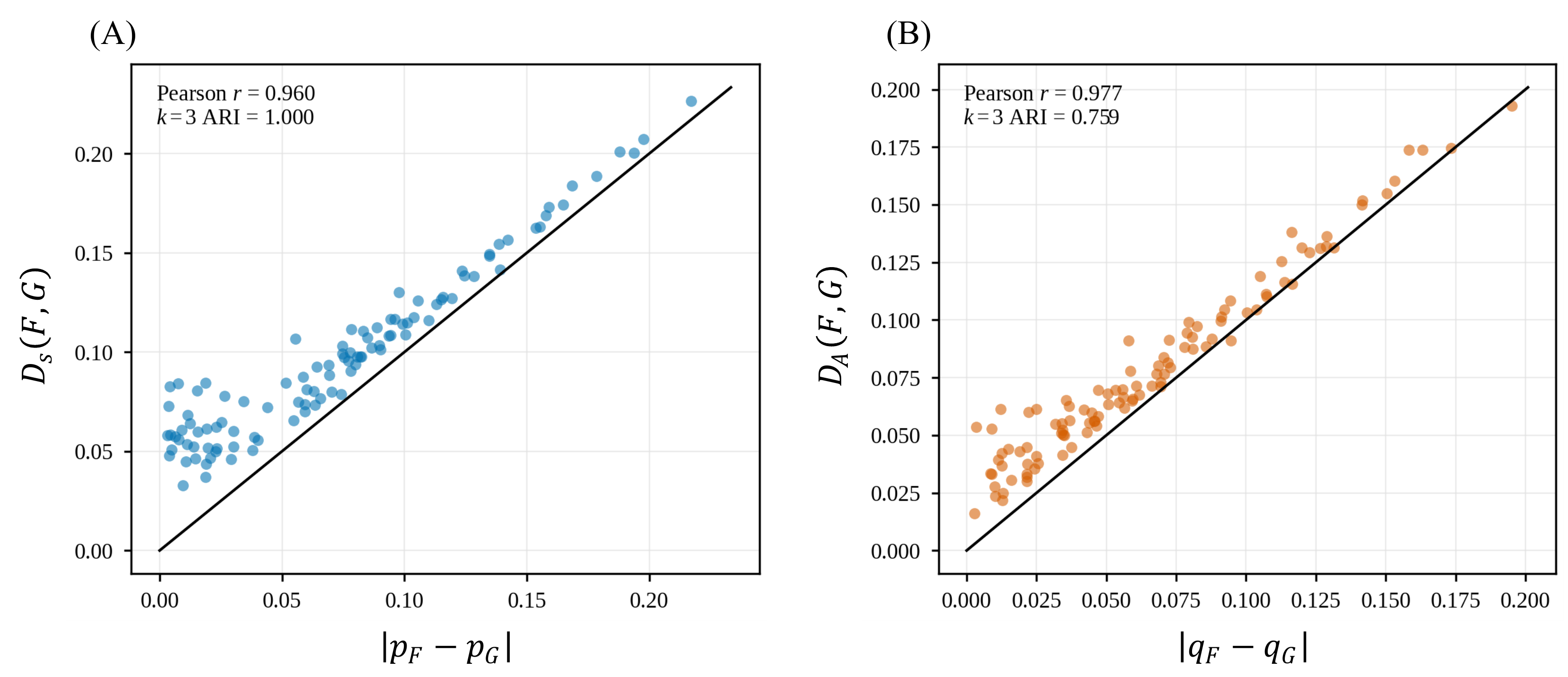}
    \caption{\textbf{Family-wide effects and profile dissimilarities.}
    Relationships between symmetric-profile dissimilarity $D_S(F,G)$ and level difference $|p_F-p_G|$, and between antisymmetric-profile dissimilarity $D_A(F,G)$ and role difference $|q_F-q_G|$, across 105 unordered family pairs in subject 01/hV4/top. Solid reference lines indicate the residual-free relationships $D_S(F,G)=|p_F-p_G|$ and $D_A(F,G)=|q_F-q_G|$.}
    \label{fig:supp_family_level_role_interpretation}
\end{figure*}

For subject 01/hV4/top, the Pearson correlation between $D_S$ and $|p_F-p_G|$ was 0.960, with a $k=3$ partition ARI of 1.000. The corresponding correlation between $D_A$ and $|q_F-q_G|$ was 0.977, with an ARI of 0.759 (Fig.~\ref{fig:supp_family_level_role_interpretation}). Across the 12 source-self conditions, the mean S--$p$ partition ARI was 0.666, and the mean A--$q$ ARI was 0.887. Because these associations are derived from the same transfer matrices, they should be interpreted descriptively. Section~\ref{sec:supp_heldout_prediction} separately evaluates the held-out-target predictive contributions of the underlying family-wide effects and pair-specific residuals.

A coarse CNN/attention-based/hybrid labeling showed low agreement with the representative $S$, $A$, and role-based $q$ partitions, with ARIs of $-0.021$, 0.067, and 0.011, respectively. Clustering based directly on negative average bidirectional transfer, $-S$, yielded a different organization dominated by a large cluster and singleton families. This comparison distinguishes direct pairwise transfer from the third-party profile similarities captured by $D_S$.

As an additional comparison, we examined representational similarity between architecture families using encoding-model responses to 200 held-out natural images in subject 01/hV4. Family-pair representational similarity correlated with symmetric transfer ($r=0.502798$ across 105 unordered family pairs; nominal $p=4.607\times10^{-8}$). The family pairs share constituent architecture families, so this nominal $p$ value is reported descriptively. In the representative condition, family-averaged encoding accuracy showed correlations of $r=0.300$ ($p=0.277$) with $p$ and $r=-0.005$ ($p=0.986$) with $q$. These associations were not consistent across source-self conditions.

\section{Pair-Specific Residuals and Robustness}
\label{sec:supp_residual_robustness}

\subsection{Residual magnitude and reliability}

For the antisymmetric component, let $\operatorname{RMS}(M)=\sqrt{105^{-1}\sum_{F<G}M_{FG}^2}$ denote the off-diagonal upper-triangle RMS within a transfer condition. Across 96 conditions, the unweighted mean of condition-wise RMS values was 0.117449 for the antisymmetric component $A$ (range: 0.050476--0.258992) and 0.033310 for the antisymmetric residual $R_A$ (range: 0.014235--0.080255). The mean condition-wise RMS ratio $\operatorname{RMS}(R_A)/\operatorname{RMS}(A)$ was 0.295895 (range: 0.181880--0.476777), indicating that the residual directional component retained approximately 30\% of the original RMS magnitude on average.

To assess reliability, component profiles were recomputed independently for each half of 10,000 repeated target splits per condition. For $S$ and $R_S$, split-half similarity was measured using Pearson correlation across 105 upper-triangle off-diagonal entries. For $A$ and $R_A$, signed cosine similarity was computed over the same entries with a fixed family ordering, equivalent to signed Frobenius cosine similarity on the full antisymmetric off-diagonal matrix. Additive effects were refitted independently within each split half before computing residual profiles. Similarities were Spearman--Brown corrected and summarized using the condition-wise median aggregation described in Section~\ref{sec:supp_transfer_reliability}.

\begin{table}[t]
    \centering
    \caption{\textbf{Component-wise split-half reliability across 96 transfer conditions.}
    Means, medians, and ranges of condition-wise median Spearman--Brown-corrected split-half similarities. Symmetric and antisymmetric components were evaluated using Pearson correlation and signed cosine similarity, respectively.}
    \label{tab:supp_component_reliability}
    \small
    \setlength{\tabcolsep}{4pt}
    \begin{tabular}{lccc}
        \toprule
        Component & Mean & Median & Range \\
        \midrule
        $S$ & 0.9811 & 0.9899 & 0.9223--0.9981 \\
        $A$ & 0.9724 & 0.9884 & 0.7843--0.9980 \\
        $R_S$ & 0.9727 & 0.9840 & 0.8728--0.9979 \\
        $R_A$ & 0.9598 & 0.9714 & 0.8569--0.9944 \\
        \bottomrule
    \end{tabular}
\end{table}

As shown in Table~\ref{tab:supp_component_reliability}, both residual components remained highly reliable despite their reduced magnitudes. Mean split-half reliability was 0.9727 for $R_S$ and 0.9598 for $R_A$, demonstrating reproducible pair-specific transfer structure across target subsets.

Tables~\ref{tab:supp_residual_conditions_s01}--\ref{tab:supp_residual_conditions_s05} report condition-wise residual statistics for all 96 transfer conditions, grouped by generating subject. For each condition, we report $\operatorname{RMS}(A)$, $\operatorname{RMS}(R_A)$, their within-condition ratio, and the split-half reliabilities of $R_S$ and $R_A$. The reported reliabilities are condition-wise medians of Spearman--Brown-corrected similarities across 10,000 target splits, as described above. These tables provide the condition-level results underlying the aggregate statistics reported in this section.

\subsection{Cross-subject residual correspondence and grouping stability}

We next examined whether pair-specific residual profiles were consistent across subjects. When the generating subject, target images, and metamers were fixed and only the evaluating subject changed, mean profile similarities across 84 comparisons were 0.888915 for $S$, 0.860599 for $A$, 0.785734 for $R_S$, and 0.673741 for $R_A$. Between native source-self conditions generated for different subjects, the corresponding mean similarities across 12 comparisons were 0.762306, 0.720293, 0.589632, and 0.461378, respectively. Thus, both symmetric and antisymmetric residual profiles retained positive cross-subject correspondence after removing family-wide effects.

We also examined the stability of residual-based family groupings. In the representative subject 01/hV4/top condition, the full-data partition agreement at $k=3$ was approximately $\mathrm{ARI}(A,R_A)=-0.002$ and $\mathrm{ARI}(S,R_A)=-0.021$, compared with an $A$--$q$ ARI of 0.759. Under target-block resampling, the residual-$A$ partition showed a median full-versus-bootstrap ARI of 1.000, with a 2.5--97.5 percentile range of [0.274, 1.000]. These results characterize the correspondence of residual profiles and the stability of their associated family partitions.

\subsection{Evaluator-model composition}

To assess sensitivity to individual evaluating models, we performed evaluator leave-one-model-out perturbations. For each perturbation, family-aggregated transfer entries were recomputed and all family-wide and residual effects were refitted, while keeping the synthesized images and target images fixed. Across source-self perturbations, the median profile similarities to the original analysis were 0.9999 for $q$ and 0.9993 for $R_A$, indicating high stability under changes in evaluator-model composition. Singleton families (AlexNet and MaxViT) were excluded from this analysis.

For directional sign sensitivity, we defined $\Delta_{FG}=T_{FG}-T_{GF}=2A_{FG}$. Among 13,104 affected pairwise comparisons involving the perturbed family, the sign of $\Delta_{FG}$ changed in 429 cases (3.27\%). The median original absolute directional difference among sign-changing comparisons was 0.011623 on the $\Delta$ scale (0.005812 on the $A$ scale). Sign reversals occurred primarily for small directional differences, with larger perturbation effects observed in some families containing two or three models.

\subsection{Alternative Response-Preservation Metrics}
\label{sec:supp_alternative_metrics}

To assess the sensitivity of directional transfer organization to the response-preservation metric, we recomputed the 12 source-self transfer matrices using three alternative measures while keeping the generated metamers, target images, and evaluating models fixed.

Let $f_m(x)$ and $z_m(x)$ denote the raw and fit-standardized predicted voxel-response vectors of evaluating model $m$, respectively, and let $V_m$ be its number of evaluated voxels. For target $x_i$ and metamer $\hat{x}_{F,i}$, we considered three higher-is-better scores:
\begin{align}
    s^{\mathrm{legacy}}_{mFi}
    &=
    -\frac{
    \|f_m(\hat{x}_{F,i})-f_m(x_i)\|_2^2
    }{
    \|f_m(x_i)\|_2^2+10^{-8}
    },\\
    s^{\mathrm{MSE}}_{mFi}
    &=
    -\frac{1}{V_m}
    \|z_m(\hat{x}_{F,i})-z_m(x_i)\|_2^2,\\
    s^{\mathrm{Pearson}}_{mFi}
    &=
    \operatorname{corr}_v
    \left(
    z_m(\hat{x}_{F,i}),z_m(x_i)
    \right).
\end{align}
The legacy NSE uses raw responses normalized by target-response energy. Response MSE uses fit-standardized responses without the reference-distance normalization of the primary metric, whereas voxel-axis Pearson measures response-pattern correlation across voxels. Pearson scores with zero centered-response norm were treated as invalid. All required scores were valid in the 12 source-self conditions.

For each metric, model-level scores were averaged equally across evaluating models within each family and then across target images, following the primary analysis. The resulting transfer matrices were decomposed using the same off-diagonal additive model as in Sec.~\ref{sec:decomposition}, yielding the family-wide directional role $q$, antisymmetric component $A$, and antisymmetric residual $R_A$.

To quantify correspondence with the primary metric, we computed Pearson correlations between the 15 family-level values of $q$. For $A$ and $R_A$, we used signed cosine similarities between their 105 upper-triangle off-diagonal entries under a fixed family ordering. These similarities retain the sign of directional relationships and are equivalent to signed Frobenius cosine similarities on the full antisymmetric matrices.

Directional sign agreement between the primary and alternative metrics was defined as
\begin{equation}
    g =
    \frac{1}{105}
    \sum_{F<G}
    \mathbf{1}
    \left[
    \operatorname{sign}(\Delta_{FG}^{\mathrm{primary}})
    =
    \operatorname{sign}(\Delta_{FG}^{\mathrm{alternative}})
    \right],
\end{equation}
where $\Delta_{FG}=T_{FG}-T_{GF}=2A_{FG}$ denotes the raw directional difference under the corresponding metric. Signs were compared without a numerical tolerance. All similarities and sign-agreement fractions were computed separately within each source-self condition and then averaged equally across the 12 conditions.

\begin{table}[!t]
    \centering
    \caption{\textbf{Correspondence of alternative response-preservation metrics with the primary transfer metric.}
    Mean profile similarities for family-wide role $q$ (Pearson correlation), antisymmetric transfer $A$ (signed cosine), and antisymmetric residual $R_A$ (signed cosine) across 12 source-self conditions. Sign agreement denotes the mean fraction of 105 unordered family pairs with matching raw directional signs.}
    \label{tab:supp_alternative_metrics}
    \small
    \setlength{\tabcolsep}{4pt}
    \renewcommand{\arraystretch}{1.12}
    \begin{tabular}{lcccc}
        \toprule
        Metric & $q$ & $A$ & $R_A$ & Sign agr. \\
        \midrule
        Legacy NSE & 0.9271 & 0.9248 & 0.9127 & 0.9095 \\
        Response MSE & 0.9830 & 0.9834 & 0.9845 & 0.9579 \\
        Voxel-axis Pearson & 0.9224 & 0.9098 & 0.7853 & 0.8714 \\
        \bottomrule
    \end{tabular}
\end{table}

As shown in Table~\ref{tab:supp_alternative_metrics}, removing reference-distance normalization (Response MSE) preserved high correspondence with the primary metric for $q$, $A$, and $R_A$ (all three mean similarities above 0.98). Legacy NSE also retained broadly similar directional organization, although correspondence was somewhat lower. Voxel-axis Pearson preserved the family-wide role ($q=0.9224$) and antisymmetric transfer ($A=0.9098$) on average, but showed lower pair-specific residual correspondence ($R_A=0.7853$), reaching 0.4754 in one condition.

These results indicate that the principal directional organization is broadly preserved across the evaluated metrics, particularly when only reference-distance normalization is removed. However, pair-specific residuals show greater metric sensitivity under some measures, especially voxel-axis Pearson. The findings therefore support descriptive robustness within the evaluated conditions.

\subsection{Target-count sensitivity}

We examined the effects of target count on component reliability at target-set sizes of 25 and 50, using 1,000 repetitions per eligible condition. When more targets were available than the requested size, each repetition sampled the requested number without replacement; when the available count equaled the requested size, the full target set was retained and only its split-half assignment was resampled. In each repetition, the selected targets were randomly divided into two disjoint, equal-sized halves: 12 and 12 for the 25-target setting, with one randomly selected target omitted, and 25 and 25 for the 50-target setting. Component profiles were obtained by averaging target-wise components within each half. Because the additive decomposition uses a fixed-design least-squares projection, the resulting residual profiles are equivalent to those obtained by fitting the additive model independently to each half-averaged transfer matrix. Split-half similarities were computed over the 105 upper-triangle off-diagonal entries using Pearson correlation for $S$ and $R_S$ and signed cosine similarity for $A$ and $R_A$. Each similarity was Spearman--Brown corrected, and the median across the 1,000 repetitions was taken as the condition-wise reliability.

Condition-wise median reliabilities were averaged equally across eligible conditions. For the 25-target setting, the means across 96 conditions were 0.974412 for $S$, 0.962600 for $A$, 0.962409 for $R_S$, and 0.943729 for $R_A$. For the 50-target setting, the corresponding means across 64 conditions were 0.991172, 0.987630, 0.986650, and 0.978588, respectively.

Matched comparisons were restricted to the same 64 transfer conditions eligible for both target counts. The mean within-condition increases in median reliability from 25 to 50 targets were 0.009309 for $S$, 0.013161 for $A$, 0.014148 for $R_S$, and 0.022196 for $R_A$. These paired improvements were computed at the condition level, whereas the preceding marginal summaries used their respective eligible condition sets. The largest improvement occurred for the antisymmetric residual $R_A$, indicating that pair-specific directional structure benefited most from additional target images.

\subsection{Shared initialization and synthesis settings}

For each generating-subject--ROI--target combination, all 15 generating families received the same initialization tensor. Before optimization, the predicted responses to this shared image depended on the evaluating family but not on the generating family, yielding a column-only initialization-stage transfer matrix. Under the additive decomposition, its antisymmetric residual was zero up to numerical precision (representative $\operatorname{RMS}(R_A)\approx2.66\times10^{-16}$).

The correlation between initialization-stage and final directional role profiles was weak (representative $r=-0.110$; pooled source-self correlation approximately 0.040). Thus, the final family-wide role structure was not simply inherited from the shared initial image.

All analyzed metamers were retained after 800 optimization iterations. Generating families contained between one and 15 backbones, resulting in different numbers of simultaneous response-preservation constraints. These synthesis settings were held fixed throughout the reported transfer analyses.

\section{Held-Out-Target Prediction}
\label{sec:supp_heldout_prediction}

\subsection{Target-wise prediction task and fold-local fitting}

To evaluate whether the decomposition captures transfer structure that generalizes beyond the targets used for fitting, we assessed predictions for held-out target images. Five-fold target cross-validation was performed separately for each of the 12 source-self generating-subject--ROI--cohort conditions. Target images were assigned to folds, preserving the complete directional family-pair profile for each training target. All family-wide effects and pair-specific residuals were estimated exclusively from the training targets of each fold.

Let $t_{i,FG}$ denote the transfer score for target $i$ and off-diagonal generating--evaluating family pair $(F,G)$, obtained by averaging the negative reference-normalized model-level errors across evaluating models in family $G$. Within each fold, we constructed a transfer matrix averaged over training targets and fitted the constrained additive model of Sec.~\ref{sec:decomposition} to its off-diagonal entries. Using the superscript ``train'' to denote effects estimated from training targets, we defined the following predictors for full transfer:
\begin{equation}
    \begin{aligned}
        M_{0,FG} &= \mu^{\mathrm{train}},\\
        M_{1,FG} &= \mu^{\mathrm{train}}+p_F^{\mathrm{train}}+p_G^{\mathrm{train}},\\
        M_{2,FG} &= \mu^{\mathrm{train}}+r_F^{\mathrm{train}}+c_G^{\mathrm{train}},\\
        M_{3,FG} &= M_{2,FG}+R_{S,FG}^{\mathrm{train}},\\
        M_{4,FG} &= M_{3,FG}+R_{A,FG}^{\mathrm{train}}.
    \end{aligned}
    \label{eq:supp_cv_transfer_models}
\end{equation}
Within each fold, the predictors assign a fixed score to each family pair across held-out targets; in particular, \(M_{4,FG}\) equals the training-target mean for that pair. This analysis therefore assesses how well family-pair transfer structure estimated from one target subset generalizes to other targets within the selected cohort. For the antisymmetric component, we defined
\begin{equation}
    \begin{aligned}
        A_{0,FG} &= 0,\\
        A_{1,FG} &= q_F^{\mathrm{train}}-q_G^{\mathrm{train}},\\
        A_{2,FG} &= A_{1,FG}+R_{A,FG}^{\mathrm{train}}.
    \end{aligned}
    \label{eq:supp_cv_direction_models}
\end{equation}
The full-transfer predictors progressively incorporate the global mean ($M_0$), family-wide levels ($M_1$), directional roles ($M_2$), symmetric residuals ($M_3$), and antisymmetric residuals ($M_4$). The antisymmetric predictors separately evaluate the contributions of directional roles ($A_1$) and pair-specific residuals ($A_2$).

For each held-out target, mean squared prediction errors were computed across 210 ordered off-diagonal family pairs for full transfer and 105 unordered family pairs for antisymmetric transfer. Errors were averaged first within each source-self condition and then equally across the 12 conditions.

\subsection{Held-out errors and incremental improvements}

\begin{table}[t]
    \centering
    \caption{\textbf{Held-out-target prediction performance.}
    Mean squared prediction errors from five-fold target cross-validation across 12 source-self conditions. All decomposition effects were fitted using only the training targets of each fold. Lower values indicate better prediction.}
    \label{tab:supp_cv_losses}
    \small
    \setlength{\tabcolsep}{4pt}
    \begin{tabular}{llr}
        \toprule
        Task & Predictor & Mean held-out MSE \\
        \midrule
        Full transfer & $M_0$ ($\mu$) & 0.08523 \\
        & $M_1$ (level) & 0.07305 \\
        & $M_2$ (level + role) & 0.06007 \\
        & $M_3$ (+ $R_S$) & 0.05755 \\
        & $M_4$ (+ $R_A$) & 0.05623 \\
        \midrule
        Antisymmetric & $A_0$ (zero) & 0.03193 \\
        & $A_1$ (role) & 0.01895 \\
        & $A_2$ (+ $R_A$) & 0.01763 \\
        \bottomrule
    \end{tabular}
\end{table}

As shown in Table~\ref{tab:supp_cv_losses}, the mean held-out MSE for full transfer decreased from 0.08523 under the global-mean predictor $M_0$ to 0.06007 under the family-wide predictor $M_2$, and further to 0.05623 after incorporating both residual components in $M_4$. Relative to $M_2$, including the residual components reduced mean MSE by 0.003847 (6.4\%). Relative to $M_0$, the full predictor $M_4$ reduced mean MSE by 0.02901 (34.0\%).

For antisymmetric transfer, mean held-out MSE decreased from 0.03193 under $A_0$ to 0.01895 under the family-wide role predictor $A_1$, and further to 0.01763 after incorporating $R_A$ in $A_2$. The incremental improvement from $A_1$ to $A_2$ was 0.001324 (7.0\%), while the total improvement from $A_0$ to $A_2$ was 0.01430 (44.8\%). Both residual-related improvements were positive in all 12 source-self conditions, demonstrating that pair-specific residuals provide additional predictive information beyond family-wide effects on held-out targets.

\subsection{Uncertainty of predictive improvements}

We assessed the uncertainty of the incremental predictive improvements using 10,000 paired target-block bootstrap replicates based on saved per-target out-of-fold losses. Within each of the 12 source-self conditions, targets were resampled with replacement using identical indices for the compared predictors. Mean MSE reductions were computed within each condition and averaged equally across conditions. Cross-validation folds, fitted predictors, and out-of-fold losses were held fixed during resampling.

The mean held-out MSE reduction from $M_2$ to $M_4$ was 0.003847, with a pointwise 95\% percentile confidence interval of [0.003501, 0.004197]. The corresponding reduction from $A_1$ to $A_2$ was 0.001324, with a pointwise 95\% confidence interval of [0.001193, 0.001459].

For the two prespecified contrasts, Bonferroni-adjusted simultaneous 95\% confidence intervals were computed using the 1.25th and 98.75th percentiles of the bootstrap distributions. The resulting intervals were [0.003453, 0.004249] for $M_2$ versus $M_4$ and [0.001175, 0.001480] for $A_1$ versus $A_2$. Both intervals excluded zero, supporting consistent improvements from incorporating pair-specific residuals.

These intervals quantify target-sampling uncertainty conditional on the selected target cohorts, fitted encoding models, fixed cross-validation folds, and out-of-fold predictions. The cross-validation evaluates generalization to unseen targets within each generating-subject--ROI--cohort condition, providing an out-of-target test of predictive structure beyond the algebraic reconstruction of the fitted transfer matrices.

\clearpage

\begin{table*}[t]
    \centering
    \caption{\textbf{Condition-wise residual statistics for generating subject 01.} Each row represents one evaluating-subject--ROI--cohort condition. The ratio is $\operatorname{RMS}(R_A)/\operatorname{RMS}(A)$; RMS values use 105 unordered off-diagonal family pairs. Reliabilities are condition-wise medians of Spearman--Brown-corrected similarities across 10,000 target splits.}
    \label{tab:supp_residual_conditions_s01}
    \footnotesize
    \setlength{\tabcolsep}{5pt}
    \renewcommand{\arraystretch}{1.05}
    \begin{tabular}{cccrrrrr}
        \toprule
        & & & \multicolumn{3}{c}{Residual magnitude} & \multicolumn{2}{c}{Split-half reliability} \\
        \cmidrule(lr){4-6}\cmidrule(lr){7-8}
        Eval. subj. & ROI & Cohort & $\operatorname{RMS}(A)$ & $\operatorname{RMS}(R_A)$ & Ratio & $R_S$ & $R_A$ \\
        \midrule
        01 & V1 & top & 0.0517 & 0.0142 & 0.2752 & 0.9964 & 0.9921 \\
        01 & V1 & bottom & 0.0862 & 0.0157 & 0.1819 & 0.9861 & 0.9755 \\
        01 & hV4 & top & 0.0823 & 0.0268 & 0.3256 & 0.9906 & 0.9834 \\
        01 & hV4 & bottom & 0.0865 & 0.0327 & 0.3776 & 0.9810 & 0.9705 \\
        \addlinespace[2pt]
        02 & V1 & top & 0.0615 & 0.0191 & 0.3101 & 0.9959 & 0.9917 \\
        02 & V1 & bottom & 0.1079 & 0.0248 & 0.2297 & 0.9837 & 0.9828 \\
        02 & hV4 & top & 0.1497 & 0.0378 & 0.2527 & 0.9920 & 0.9859 \\
        02 & hV4 & bottom & 0.0964 & 0.0288 & 0.2992 & 0.9589 & 0.9505 \\
        \addlinespace[2pt]
        03 & V1 & top & 0.0505 & 0.0241 & 0.4768 & 0.9939 & 0.9926 \\
        03 & V1 & bottom & 0.1769 & 0.0362 & 0.2048 & 0.9872 & 0.9838 \\
        03 & hV4 & top & 0.1261 & 0.0388 & 0.3074 & 0.9913 & 0.9843 \\
        03 & hV4 & bottom & 0.1008 & 0.0366 & 0.3628 & 0.9738 & 0.9643 \\
        \addlinespace[2pt]
        04 & V1 & top & 0.0772 & 0.0245 & 0.3175 & 0.9965 & 0.9934 \\
        04 & V1 & bottom & 0.1008 & 0.0219 & 0.2174 & 0.9814 & 0.9753 \\
        04 & hV4 & top & 0.1428 & 0.0368 & 0.2579 & 0.9927 & 0.9848 \\
        04 & hV4 & bottom & 0.0794 & 0.0235 & 0.2957 & 0.9736 & 0.9519 \\
        \addlinespace[2pt]
        05 & V1 & top & 0.1360 & 0.0399 & 0.2933 & 0.9970 & 0.9934 \\
        05 & V1 & bottom & 0.0969 & 0.0313 & 0.3233 & 0.9707 & 0.9773 \\
        05 & hV4 & top & 0.1640 & 0.0452 & 0.2754 & 0.9925 & 0.9835 \\
        05 & hV4 & bottom & 0.0680 & 0.0252 & 0.3700 & 0.9636 & 0.9328 \\
        \addlinespace[2pt]
        06 & V1 & top & 0.1217 & 0.0258 & 0.2117 & 0.9979 & 0.9932 \\
        06 & V1 & bottom & 0.0579 & 0.0161 & 0.2782 & 0.9716 & 0.9707 \\
        06 & hV4 & top & 0.2149 & 0.0642 & 0.2988 & 0.9949 & 0.9918 \\
        06 & hV4 & bottom & 0.0847 & 0.0279 & 0.3296 & 0.9718 & 0.9552 \\
        \addlinespace[2pt]
        07 & V1 & top & 0.0517 & 0.0188 & 0.3630 & 0.9923 & 0.9893 \\
        07 & V1 & bottom & 0.1563 & 0.0332 & 0.2125 & 0.9843 & 0.9842 \\
        07 & hV4 & top & 0.0741 & 0.0309 & 0.4165 & 0.9872 & 0.9827 \\
        07 & hV4 & bottom & 0.1116 & 0.0348 & 0.3114 & 0.9757 & 0.9668 \\
        \addlinespace[2pt]
        08 & V1 & top & 0.0913 & 0.0287 & 0.3138 & 0.9969 & 0.9944 \\
        08 & V1 & bottom & 0.1112 & 0.0265 & 0.2383 & 0.9819 & 0.9810 \\
        08 & hV4 & top & 0.1902 & 0.0480 & 0.2525 & 0.9959 & 0.9928 \\
        08 & hV4 & bottom & 0.0844 & 0.0260 & 0.3077 & 0.9655 & 0.9619 \\
        \bottomrule
    \end{tabular}
\end{table*}

\clearpage

\begin{table*}[t]
    \centering
    \caption{\textbf{Condition-wise residual statistics for generating subject 02.} Each row represents one evaluating-subject--ROI--cohort condition. The ratio is $\operatorname{RMS}(R_A)/\operatorname{RMS}(A)$; RMS values use 105 unordered off-diagonal family pairs. Reliabilities are condition-wise medians of Spearman--Brown-corrected similarities across 10,000 target splits.}
    \label{tab:supp_residual_conditions_s02}
    \footnotesize
    \setlength{\tabcolsep}{5pt}
    \renewcommand{\arraystretch}{1.05}
    \begin{tabular}{cccrrrrr}
        \toprule
        & & & \multicolumn{3}{c}{Residual magnitude} & \multicolumn{2}{c}{Split-half reliability} \\
        \cmidrule(lr){4-6}\cmidrule(lr){7-8}
        Eval. subj. & ROI & Cohort & $\operatorname{RMS}(A)$ & $\operatorname{RMS}(R_A)$ & Ratio & $R_S$ & $R_A$ \\
        \midrule
        01 & V1 & top & 0.0743 & 0.0222 & 0.2991 & 0.9958 & 0.9935 \\
        01 & V1 & bottom & 0.1174 & 0.0220 & 0.1875 & 0.9741 & 0.9654 \\
        01 & hV4 & top & 0.0950 & 0.0279 & 0.2936 & 0.9853 & 0.9694 \\
        01 & hV4 & bottom & 0.0921 & 0.0360 & 0.3906 & 0.9602 & 0.9497 \\
        \addlinespace[2pt]
        02 & V1 & top & 0.0939 & 0.0273 & 0.2903 & 0.9970 & 0.9928 \\
        02 & V1 & bottom & 0.1174 & 0.0220 & 0.1875 & 0.9817 & 0.9578 \\
        02 & hV4 & top & 0.1747 & 0.0423 & 0.2422 & 0.9874 & 0.9765 \\
        02 & hV4 & bottom & 0.0990 & 0.0358 & 0.3618 & 0.9453 & 0.9479 \\
        \addlinespace[2pt]
        03 & V1 & top & 0.1093 & 0.0399 & 0.3648 & 0.9960 & 0.9944 \\
        03 & V1 & bottom & 0.2060 & 0.0400 & 0.1943 & 0.9820 & 0.9700 \\
        03 & hV4 & top & 0.1524 & 0.0434 & 0.2848 & 0.9822 & 0.9659 \\
        03 & hV4 & bottom & 0.1169 & 0.0410 & 0.3511 & 0.9557 & 0.9434 \\
        \addlinespace[2pt]
        04 & V1 & top & 0.1268 & 0.0356 & 0.2810 & 0.9965 & 0.9934 \\
        04 & V1 & bottom & 0.1131 & 0.0225 & 0.1988 & 0.9764 & 0.9527 \\
        04 & hV4 & top & 0.1564 & 0.0393 & 0.2510 & 0.9866 & 0.9732 \\
        04 & hV4 & bottom & 0.0710 & 0.0214 & 0.3010 & 0.9351 & 0.8961 \\
        \addlinespace[2pt]
        05 & V1 & top & 0.2046 & 0.0620 & 0.3032 & 0.9959 & 0.9935 \\
        05 & V1 & bottom & 0.1356 & 0.0347 & 0.2556 & 0.9663 & 0.9508 \\
        05 & hV4 & top & 0.1823 & 0.0498 & 0.2730 & 0.9859 & 0.9700 \\
        05 & hV4 & bottom & 0.0543 & 0.0230 & 0.4230 & 0.9182 & 0.8727 \\
        \addlinespace[2pt]
        06 & V1 & top & 0.1810 & 0.0370 & 0.2046 & 0.9973 & 0.9920 \\
        06 & V1 & bottom & 0.0762 & 0.0172 & 0.2250 & 0.9659 & 0.9368 \\
        06 & hV4 & top & 0.2355 & 0.0715 & 0.3035 & 0.9908 & 0.9860 \\
        06 & hV4 & bottom & 0.0642 & 0.0236 & 0.3670 & 0.9172 & 0.8746 \\
        \addlinespace[2pt]
        07 & V1 & top & 0.0629 & 0.0236 & 0.3749 & 0.9949 & 0.9891 \\
        07 & V1 & bottom & 0.1658 & 0.0313 & 0.1889 & 0.9715 & 0.9636 \\
        07 & hV4 & top & 0.0738 & 0.0330 & 0.4473 & 0.9763 & 0.9654 \\
        07 & hV4 & bottom & 0.1142 & 0.0377 & 0.3298 & 0.9514 & 0.9427 \\
        \addlinespace[2pt]
        08 & V1 & top & 0.1645 & 0.0440 & 0.2676 & 0.9970 & 0.9942 \\
        08 & V1 & bottom & 0.1394 & 0.0261 & 0.1874 & 0.9756 & 0.9527 \\
        08 & hV4 & top & 0.2156 & 0.0539 & 0.2502 & 0.9924 & 0.9875 \\
        08 & hV4 & bottom & 0.0994 & 0.0306 & 0.3077 & 0.9280 & 0.9361 \\
        \bottomrule
    \end{tabular}
\end{table*}

\clearpage

\begin{table*}[t]
    \centering
    \caption{\textbf{Condition-wise residual statistics for generating subject 05.} Each row represents one evaluating-subject--ROI--cohort condition. The ratio is $\operatorname{RMS}(R_A)/\operatorname{RMS}(A)$; RMS values use 105 unordered off-diagonal family pairs. Reliabilities are condition-wise medians of Spearman--Brown-corrected similarities across 10,000 target splits.}
    \label{tab:supp_residual_conditions_s05}
    \footnotesize
    \setlength{\tabcolsep}{5pt}
    \renewcommand{\arraystretch}{1.05}
    \begin{tabular}{cccrrrrr}
        \toprule
        & & & \multicolumn{3}{c}{Residual magnitude} & \multicolumn{2}{c}{Split-half reliability} \\
        \cmidrule(lr){4-6}\cmidrule(lr){7-8}
        Eval. subj. & ROI & Cohort & $\operatorname{RMS}(A)$ & $\operatorname{RMS}(R_A)$ & Ratio & $R_S$ & $R_A$ \\
        \midrule
        01 & V1 & top & 0.0653 & 0.0222 & 0.3398 & 0.9951 & 0.9934 \\
        01 & V1 & bottom & 0.0706 & 0.0158 & 0.2241 & 0.9448 & 0.9123 \\
        01 & hV4 & top & 0.1116 & 0.0289 & 0.2593 & 0.9858 & 0.9658 \\
        01 & hV4 & bottom & 0.0860 & 0.0381 & 0.4433 & 0.9146 & 0.9145 \\
        \addlinespace[2pt]
        02 & V1 & top & 0.0867 & 0.0256 & 0.2957 & 0.9960 & 0.9909 \\
        02 & V1 & bottom & 0.0664 & 0.0175 & 0.2630 & 0.9555 & 0.9208 \\
        02 & hV4 & top & 0.1854 & 0.0429 & 0.2315 & 0.9887 & 0.9721 \\
        02 & hV4 & bottom & 0.0977 & 0.0361 & 0.3694 & 0.8728 & 0.8913 \\
        \addlinespace[2pt]
        03 & V1 & top & 0.0913 & 0.0373 & 0.4087 & 0.9948 & 0.9936 \\
        03 & V1 & bottom & 0.1139 & 0.0242 & 0.2123 & 0.9406 & 0.9039 \\
        03 & hV4 & top & 0.1942 & 0.0500 & 0.2577 & 0.9857 & 0.9707 \\
        03 & hV4 & bottom & 0.0891 & 0.0360 & 0.4048 & 0.9004 & 0.8909 \\
        \addlinespace[2pt]
        04 & V1 & top & 0.1203 & 0.0338 & 0.2812 & 0.9961 & 0.9923 \\
        04 & V1 & bottom & 0.0579 & 0.0174 & 0.3008 & 0.9430 & 0.9114 \\
        04 & hV4 & top & 0.1830 & 0.0471 & 0.2575 & 0.9893 & 0.9771 \\
        04 & hV4 & bottom & 0.0781 & 0.0270 & 0.3464 & 0.9318 & 0.8872 \\
        \addlinespace[2pt]
        05 & V1 & top & 0.2005 & 0.0635 & 0.3170 & 0.9962 & 0.9943 \\
        05 & V1 & bottom & 0.0635 & 0.0217 & 0.3413 & 0.9223 & 0.8757 \\
        05 & hV4 & top & 0.2191 & 0.0728 & 0.3325 & 0.9877 & 0.9842 \\
        05 & hV4 & bottom & 0.0752 & 0.0307 & 0.4078 & 0.9344 & 0.8569 \\
        \addlinespace[2pt]
        06 & V1 & top & 0.1751 & 0.0364 & 0.2078 & 0.9972 & 0.9929 \\
        06 & V1 & bottom & 0.0616 & 0.0150 & 0.2435 & 0.9602 & 0.9190 \\
        06 & hV4 & top & 0.2590 & 0.0803 & 0.3099 & 0.9929 & 0.9875 \\
        06 & hV4 & bottom & 0.1134 & 0.0348 & 0.3067 & 0.9356 & 0.8972 \\
        \addlinespace[2pt]
        07 & V1 & top & 0.0573 & 0.0250 & 0.4366 & 0.9935 & 0.9886 \\
        07 & V1 & bottom & 0.1051 & 0.0228 & 0.2167 & 0.9301 & 0.9228 \\
        07 & hV4 & top & 0.1172 & 0.0317 & 0.2704 & 0.9761 & 0.9595 \\
        07 & hV4 & bottom & 0.0978 & 0.0302 & 0.3083 & 0.9067 & 0.8658 \\
        \addlinespace[2pt]
        08 & V1 & top & 0.1605 & 0.0422 & 0.2629 & 0.9970 & 0.9938 \\
        08 & V1 & bottom & 0.0651 & 0.0179 & 0.2742 & 0.9352 & 0.8836 \\
        08 & hV4 & top & 0.2444 & 0.0604 & 0.2472 & 0.9939 & 0.9875 \\
        08 & hV4 & bottom & 0.0859 & 0.0302 & 0.3516 & 0.9067 & 0.8756 \\
        \bottomrule
    \end{tabular}
\end{table*}


\end{document}